\documentclass[11pt]{article}

\usepackage[table]{xcolor}
\PassOptionsToPackage{hyphens}{url}

\usepackage{booktabs}
\usepackage{multirow}
\usepackage{pifont}
\usepackage{caption}
\usepackage{subcaption}
\usepackage{xspace}
\usepackage{microtype}
\usepackage{tikz}                 % corner Qualcomm mark
\usepackage[numbers]{natbib}      % must precede hyperref

\usepackage[margin=1in, top=1in]{geometry}
\usepackage{graphicx}
\usepackage{fancyhdr}
\usepackage{hyperref}
\usepackage{xcolor}
\usepackage{titlesec}
\usepackage{tcolorbox}
\usepackage{titling}
\tcbuselibrary{skins}

\usepackage[T1]{fontenc}
\usepackage{tgheros}
\usepackage[utf8]{inputenc}

\usepackage{amsmath,amssymb,bm}
\usepackage{newtxtext}
\usepackage{newtxmath} % reuses AMS alphabets to avoid overflow

\AtBeginDocument{%

}

\providecommand{\titlefont}{\sffamily\bfseries}
\providecolor{qc_darkblue}{rgb}{0.008,0.063,0.247}
\providecolor{qc_blue}{rgb}{0.164,0.164,0.914}

\titleformat{\title}
{\titlefont\LARGE\color{qc_blue}}{}{0pt}{}

\titleformat{\section}
{\titlefont\Large\bfseries\color{qc_darkblue}}{\thesection}{1em}{}

\titlespacing*{\section}{0em}{1em}{.6em}

\newtcolorbox{titlebox}{
  enhanced,
  colback=white,
  boxrule=0pt,
  opacityback=0,
  opacityframe=0,
  width=0.95\textwidth,
  center
}

\makeatletter
\newcommand{\contactinfo}[1]{\def\@contactinfo{#1}}
\makeatother
\contactinfo{}

\fancypagestyle{titlepage}{
  \fancyhf{}
  \fancyfoot[L]{\footnotesize Qualcomm AI Research is an initiative of Qualcomm Technologies, Inc.}
  \fancyfoot[R]{\footnotesize\thepage}
  
}
\renewcommand\abstract{%
    \setlength{\parskip}{.8em}
    \par
}

\title{Paper Title}
\date{February 23, 2024}
\author{Author1, Author2}
\contactinfo{\{author1, author2\}@qualcomm.com}

\usepackage{mathtools}            % \mathclap, used in src/method.tex
\usepackage{cleveref}             % no options: preserves the existing
\usepackage{array}
\newcolumntype{X}{l}
\newsavebox{\squadtabbox}
\newenvironment{squadautotab}[1]
  {\begin{lrbox}{\squadtabbox}\begin{tabular}{#1}}
  {\end{tabular}\end{lrbox}%
   \ifdim\wd\squadtabbox>\linewidth
     \resizebox{\linewidth}{!}{\usebox{\squadtabbox}}%
   \else
     \usebox{\squadtabbox}%
   \fi}
\newenvironment{tabularx}[2]{\begin{squadautotab}{#2}}{\end{squadautotab}}
\renewenvironment{tabular*}[2]{\begin{squadautotab}{#2}}{\end{squadautotab}}

\providecommand{\Real}{\mathbb{R}}
\providecommand{\bigO}{\mathcal{O}}

\providecommand{\E}{\mathbb{E}}
\providecommand{\normal}{\mathcal{N}}
\providecommand{\loss}{\mathcal{L}}
\DeclareMathOperator*{\softmax}{softmax}

\DeclareMathOperator{\Attn}{Attn}
\newcommand{\AttnL}{\operatorname{Attn}\!\mathcal{L}}   % local  attention  Attn��
\newcommand{\AttnG}{\operatorname{Attn}\!\mathcal{G}}   % global attention  Attn��

\newcommand{\loc}{l}                                    % local  superscript
\newcommand{\glob}{g}                                   % global superscript

\providecommand{\bX}{\mathbf{X}}

\providecommand{\bY}{\mathbf{Y}}
\providecommand{\bQ}{\mathbf{Q}}
\providecommand{\bK}{\mathbf{K}}
\providecommand{\bV}{\mathbf{V}}
\providecommand{\bW}{\mathbf{W}}

\providecommand{\bI}{\mathbf{I}}

\providecommand{\zero}{\mathbf{0}}

\DeclareMathOperator{\SQuadAttn}{SQuadAttn}
\DeclareMathOperator{\SelfAttn}{SelfAttn}

\DeclareMathOperator{\splitL}{split\mathcal{L}}
\DeclareMathOperator{\splitG}{split\mathcal{G}}
\newcommand{\name}{SQuad\xspace}
\newcommand{\namef}{Sub-Quadratic Attention Distillation\xspace}

\makeatletter
\renewenvironment{figure*}{\@float{figure}}{\end@float}
\renewenvironment{table*}{\@float{table}}{\end@float}
\makeatother

\renewcommand{\twocolumn}[1][]{\newpage #1}

\newcommand{\citesupp}[1]{\cite{#1}}
\newcommand{\bibliographystylesupp}[1]{}
\newcommand{\bibliographysupp}[1]{}

\renewenvironment{abstract}
  {\setlength{\parskip}{.8em}\par}
  {\par}

\newif\ifsupp
\supptrue
\ifsupp
  \newcommand{\suppref}[1]{\cref{#1}}
\else
  \newcommand{\suppref}[1]{the supplementary material}
\fi

\hypersetup{
  pdftitle={SQuad: Sub-Quadratic Attention Distillation for Efficient Video Generation},
  pdfauthor={Animesh Karnewar, Denis Korzhenkov, Amirhossein Habibian, Mohsen Ghafoorian},
  colorlinks=true,
  linkcolor=qc_darkblue,
  citecolor=qc_darkblue,
  urlcolor=qc_blue,
}

\begin{document}
\thispagestyle{titlepage}
\bibliographystyle{unsrtnat}   % numeric, natbib-aware: keeps \citet working

% ---- Radial Qualcomm mark bleeding off the top-right corner ---------
\begin{tikzpicture}[remember picture,overlay]
  \node[anchor=north east, inner sep=0pt]
    at ([xshift=5cm,yshift=5.5cm]current page.north east)
    {\includegraphics[width=10cm,angle=-75,origin=c]%
      {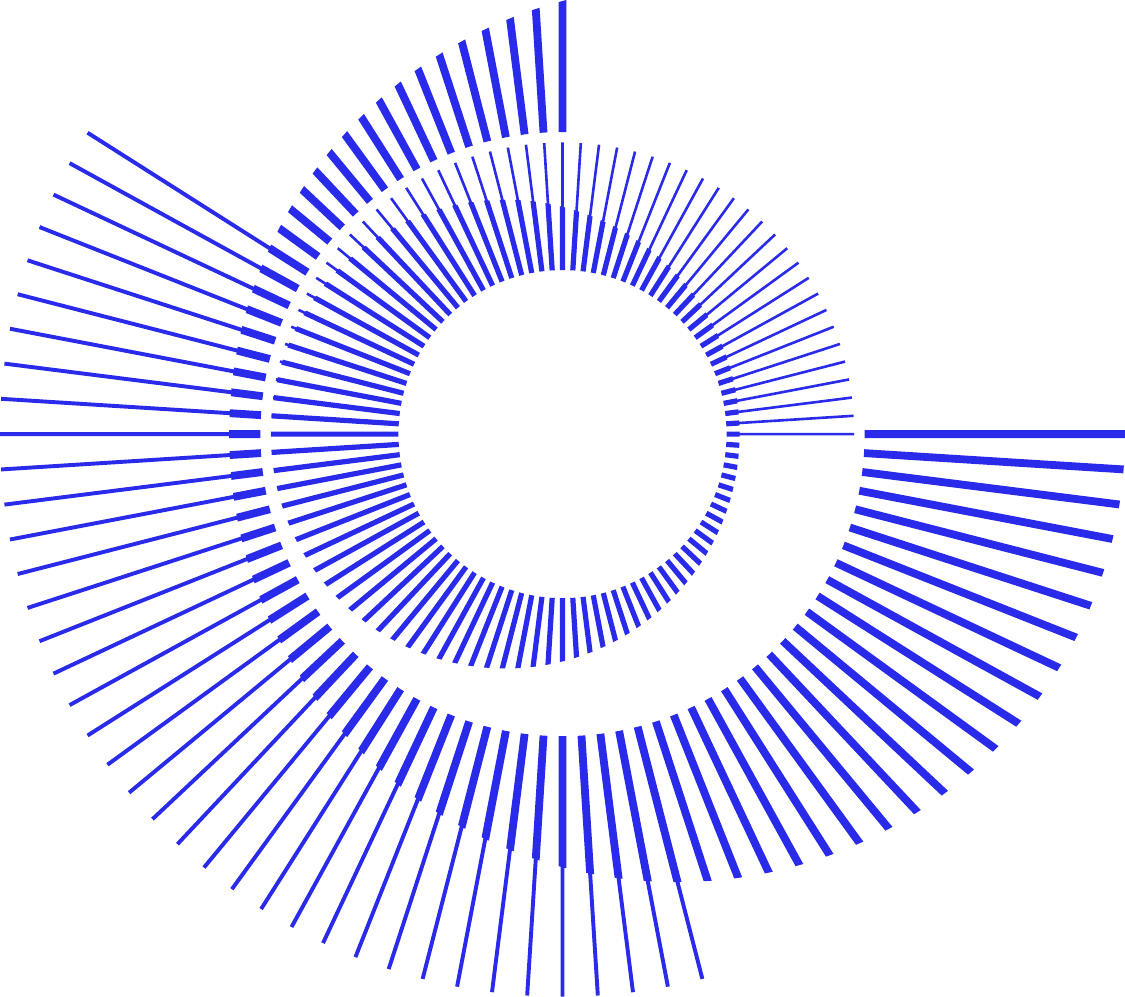}};
\end{tikzpicture}

% ---- Qualcomm AI Research lockup ------------------------------------
\begin{figure}[t]
    \vspace*{-1cm}
    \hspace*{-0.6cm}
    \includegraphics[width=4.0cm]{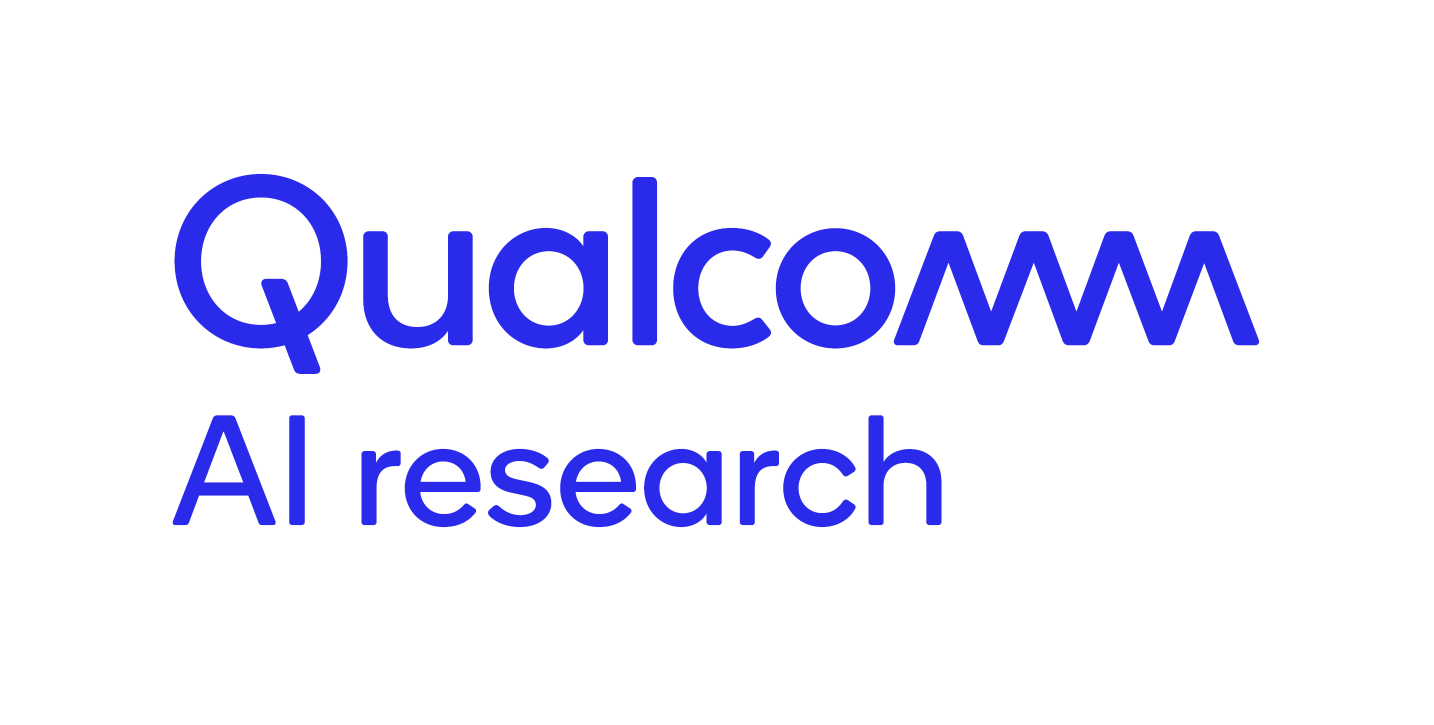}
    \vspace*{-0.5cm}
\end{figure}
\vspace{-2em}

% ---- Title / authors -------------------------------------------------
% Replaces \input{src/title.tex}: the AAAI \affiliations macro does not
% exist outside aaai2027.sty, and the \thanks footnote it carried is
% already in the page footer supplied by format_air_v3.
\title{\textbf{\name}: \LARGE{\namef for Efficient Video Generation}}
\date{August 3, 2026}
\author{
    Animesh Karnewar,
    Denis Korzhenkov,
    Amirhossein Habibian,
    Mohsen Ghafoorian
}
\contactinfo{\{karnewar, dkorzhen, ahabibia, mghafoor\}@qualcomm.com}

\begin{titlebox}
{\titlefont\huge\bfseries\color{qc_darkblue}\thetitle}\\

\makeatletter
{\titlefont\mdseries\normalfont\small\color{qc_blue}\@author\par
 \vspace{0.3em}
 % Qualcomm AI Research\par
 % Matrix ONE, Science Park 301, 1098 XH Amsterdam, Netherlands\par
 % \ifx\@contactinfo\@empty
 % \else
   % {\bfseries\@contactinfo}%
 % \fi
}
\makeatother

% ---- Teaser -----------------------------------------------------------
% In the AAAI build this was injected by patching \@maketitle so it
% could not float away from the title. Here the title block is an
% ordinary tcolorbox, so the figure simply sits inside it. \linewidth is
% the box interior (titlebox is 0.95\textwidth wide), not the page.
\begin{center}
  \includegraphics[width=\linewidth, trim=0 5 0 18, clip]{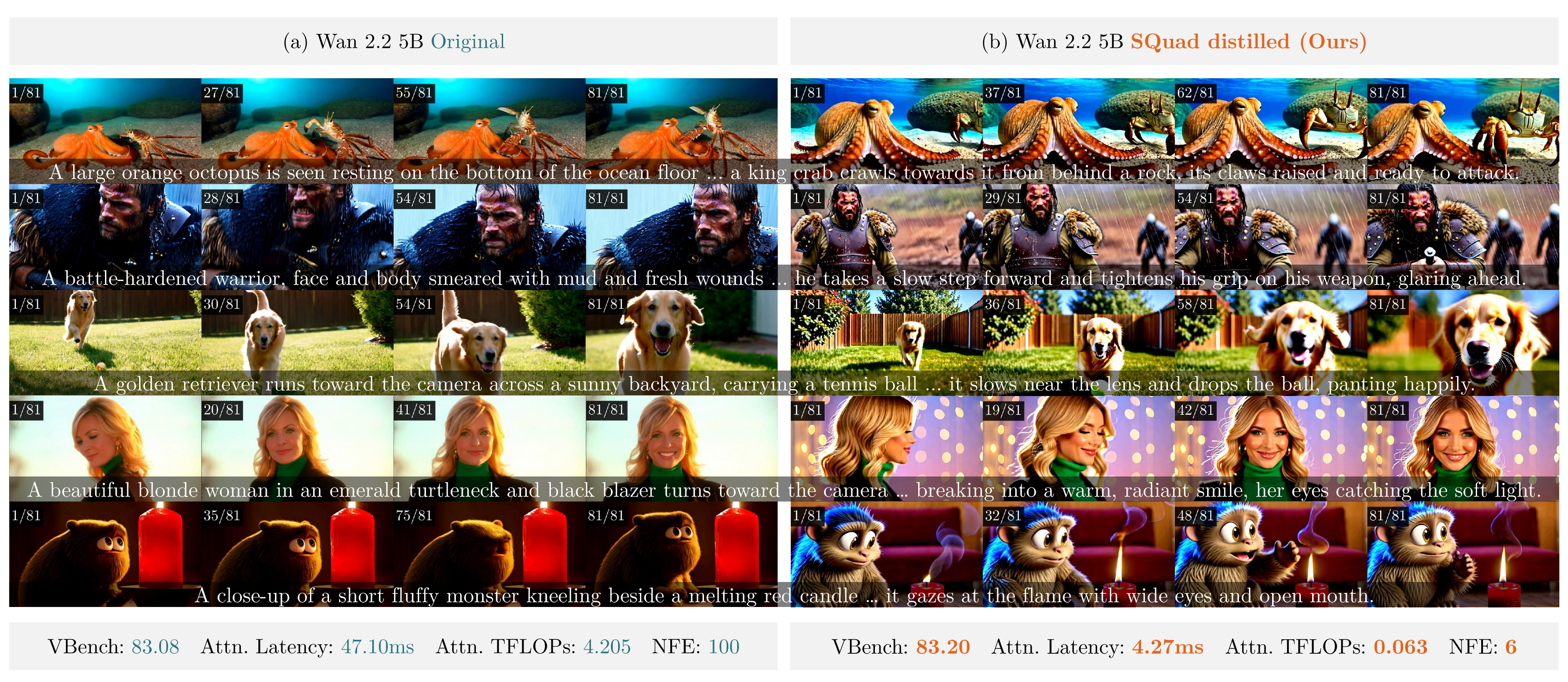}
  \captionof{figure}{\textbf{Efficient video generation with \name.} Given
    the same prompts, the original Wan 2.2 5B model (a) produces
    high-quality videos but at significant computational cost. Our
    \name-distilled model (b) preserves the quality while substantially
    reducing the computational cost.}
  \label{fig:teaser}
\end{center}

% src/abstract.tex already contains \begin{abstract} ... \end{abstract}
\begin{abstract}

Video Diffusion Transformers (DiTs) spend most of their compute inside the 
Self-Attention operation, whose cost grows quadratically, $\mathcal{O}(n^2)$, with
the number of latent tokens $n$. For the task of video generation, the token count is large, so this
term dominates runtime and memory, and thereby caps the resolution and duration
we can generate. Linear $\mathcal{O}(n)$ and low-rank $\mathcal{O}(nk)$ surrogates of Self-Attention trade the full softmax $QK^T$  for cheaper kernels, but rarely recover the original's expressivity, leaving a
stubborn quality gap. Motivated by this, we propose \name, a \namef framework that achieves a complexity of $\mathcal{O}(n\sqrt{n})$ in the resulting distilled Attention, naturally balancing the efficiency v/s expressivity trade-off. Instead of training our own Video DiT from scratch, which is prohibitively expensive, we fit a pretrained full softmax Self-Attention DiT into our proposed SQuad-Attention one by distilling the former in two stages: Flow-Matching
Supervised Fine-Tuning (SFT), followed by improved Distribution Matching Distillation (DMD2) which additionally makes the sampling more efficient. On the Wan~2.2 5B
text-to-video model, SQuAD matches the quadratic teacher on VBench
($83.20$ v/s $83.08$) while cutting the per-step per-block attention FLOPs by $\sim$$67\times$ and attention latency by $\sim$$11\times$, and end-to-end DiT latency by 2$\times$, all while also generating a video in only $6$ Neural Functional Evaluations (NFEs) instead of the default $100$.

\end{abstract}

\end{titlebox}

% ---- Main body -------------------------------------------------------
\section{Introduction}

% Idea notes: 
% \begin{enumerate}
%     \item Para 1: Introduce Video Generation, and show how important SoftMax Attention is for Video Generation. 
%     \item Para 2: Challenging because of the quadratic complexity and Linear cheapness not achieving the expressivity required. 
%     \item Para 3: Prior solutions with Hybridization of the Transformer Model. 
%     \item Para 4: We introduce SQuAD, and how it solves all the problems in the world. Transition into Contributions.
% \end{enumerate}

% This is the method figure - part 1,  
% !!placing here for better readability!!
\begin{figure*}[t]
\centering
\includegraphics[width=\textwidth]{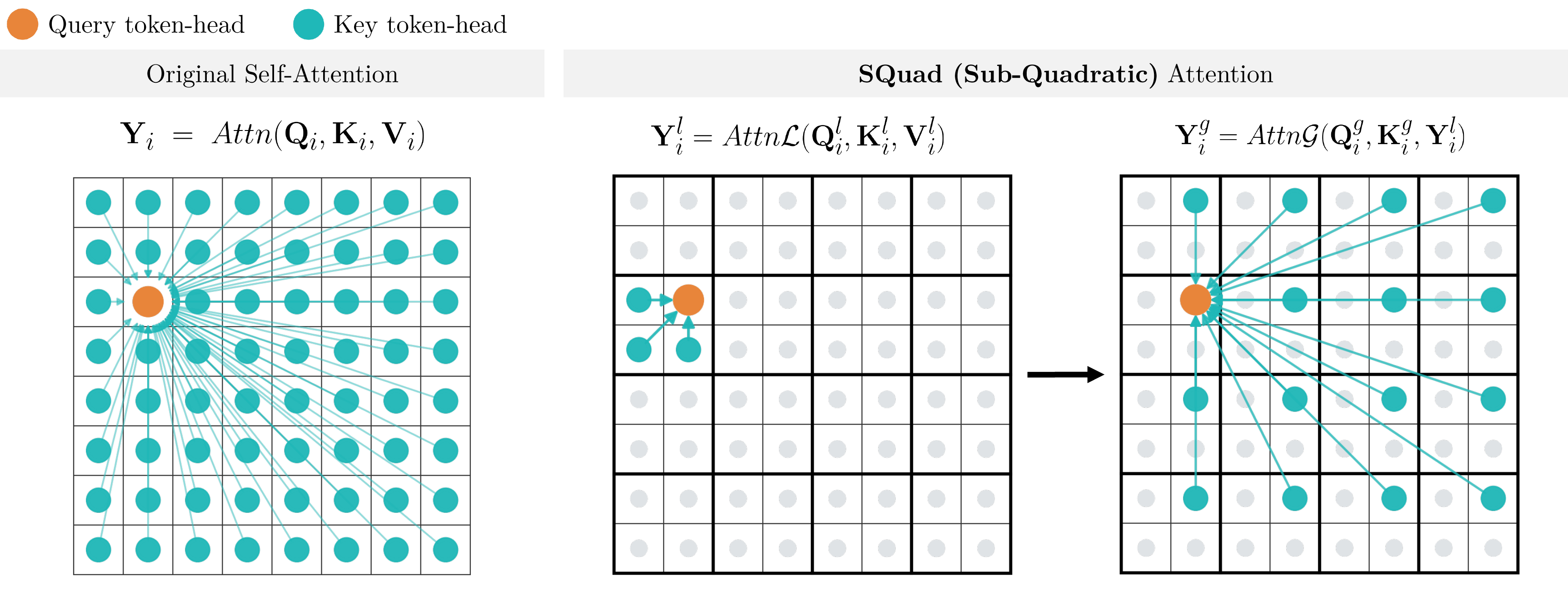} 
\caption{%
  \textbf{\name{} attention.} \emph{Left:} full softmax Self-Attention at
  $\mathcal{O}(n^2)$ complexity. \emph{Right:} \name{} factorizes it into a \emph{local}
  pass \emph{within} $\mathcal{O}(\sqrt{n})$ windows followed by a \emph{global} pass
  \emph{across} the windows, giving a full receptive field at
  $\mathcal{O}(n\sqrt{n})$ complexity with a true softmax throughout.%
}
\label{fig:method_part1}
\end{figure*}

Video generation has quickly become one of the most exciting frontiers of
generative modeling. DiTs~\cite{peebles2023dit} now
produce high-resolution, temporally coherent clips of remarkable visual
quality~\cite{brooks2024sora,blattmann2023svd,wan2025}. On today's hardware,
however, these models generate only a few seconds of video in a single
pass~\cite{wan2025}. Longer videos are stitched together by generating short
clips one after another, each conditioned on the last---a process that ultimately
drifts and loses coherence as the video grows~\cite{li2026stablevideoinfinity, cui2026selfforcingpp, liu2026streaming, chen2026sanavideo, huang2026steinsgate, yang2026longlive}. 
%liu2026rollingforcing, cai2026mixtureofcontexts,
Generating longer videos with higher resolution natively,
therefore remains a central goal of the field. Underlying all of these models is
the softmax Self-Attention mechanism~\cite{vaswani2017attention}, which lets
every token attend to every other token and is widely seen as the main driver of
the expressivity and scalability of modern transformers. Softmax Self-Attention
matters far beyond video: it is the backbone of nearly every domain that
transformers now dominate. Studying how our proposed modification generalizes
across all of these settings would be fascinating, but in favor of a concrete and carefully
evaluated research, we focus on video generation with a well-defined scope, where the cost of attention is the highest and the potential payoff is perhaps one of the
greatest, if not the greatest.

The biggest challenge is computational cost. Softmax Self-Attention scales quadratically with the number of
tokens, $\mathcal{O}(n^2)$, in both time and memory~\cite{vaswani2017attention}.
In video, a single token indexes a point in space and time, and their number
easily reaches tens of thousands per frame-volume. At this scale the
quadratic term dominates the entire generation budget and caps the resolution,
duration, and throughput the GPU hardware can afford. The obvious fix is to make the attention
cheaper. Linear~$\mathcal{O}(n)$ attention and related 
$\mathcal{O}(nk)$~approximations~\cite{katharopoulos2020linear,choromanski2021performer,%
wang2020linformer} replace the softmax with a kernelized or low-rank surrogate
and bring the cost down. But these surrogates rarely match the expressivity of the full
softmax Self-Attention. The non-linearity and the sharp, input-dependent selectivity
that make softmax Self-Attention so effective are exactly what the approximations
give up. The result is a stubborn quality gap, and it is widest in the
high-fidelity, detail-sensitive regime of video generation.

Faced with this trade-off, prior works have looked for a middle ground. Whether it is
linearizing a pretrained model or distilling its attention into
kernel-based approximations~\cite{katharopoulos2020linear,choromanski2021performer}, or even
replacing the attention altogether with structured state-space models such as
S4 and Mamba~\cite{gu2022s4,gu2023mamba}, on their own, neither substitution quite matches the performance and scalability of
full softmax Self-Attention. Hence the field has increasingly turned to \emph{hybrid} designs
that interleave a few expensive quadratic full softmax Self-Attention blocks with many cheaper
efficient ones~\cite{lieber2024jamba}. This idea has recently reached video
generation as well~\cite{ghafoorian2025attentionsurgery,ghafoorian2026rehyat,zhang2025vsa,li2025radial,yang2025svg2,zhang2025jenga,ghafoorian2026mobilewan}.
These hybrids recover much of the lost quality, yet they pay for it with complex-heterogeneous
architectures, delicate choices about which layers stay quadratic, and more-exacerbatingly, a lingering
dependence on the very softmax Self-Attention operation they set out to replace.

Considering this scenario, we take a different route with \textbf{\name}, a \emph{\namef} framework for efficient video generation. Rather than abandoning
softmax Self-Attention, we lean into one of its well-known properties: in video DiTs,
the attention maps are sparse and heavy-tailed. Almost all of the attention mass
falls on a small set of critical tokens, while the rest of the
$\mathrm{softmax}(QK^\top/\sqrt{d})$ entries stay close to
zero~\cite{zhang2025vsa,xi2025svg,svg2,sparsevdit}. We turn this observation
into a framework that lowers the attention cost from
$\mathcal{O}(n^2)$ to $\mathcal{O}(n\sqrt{n})$ while keeping a genuine softmax
throughout, unlike the linear variants (glance over \cref{fig:method_part1}). Our proposed version of 
attention is rather simple but surprisingly effective. 
We propose a fixed, reduced communication pattern with two steps:
locally, tokens mix within $\mathcal{O}(\sqrt{n})$-sized windows;
while globally, tokens at the same position in each window mix across all windows.
On the Wan~2.2 5B text-to-video generation model, as well as 2.1 1.3B, \name distills the 
quadratic full softmax Self-Attention into our proposed sub-quadratic \name-Attention with almost no loss in
quality, while cutting latency and TFLOPs substantially (see \cref{fig:teaser}).

\section{Related Work}

We cover the more relevant related works on efficient attention methods (\cref{sec:relwork:eff_attn}), 
and Distillation of Video Generators (\cref{sec:relwork:distillation}) here, while defer 
the broader coverage on Text-to-Video generation models to the supplementary (\suppref{sec:supp_relwork}).

\subsection{Efficient Attention /  Token Merging}
\label{sec:relwork:eff_attn}
Because this is such an important problem, a large body
of work has sought to solve it. One line replaces or approximates
the softmax. Attention~Surgery~\cite{ghafoorian2025attentionsurgery}
linearizes a pretrained video diffusion transformer by distilling its
softmax attention into a hybrid softmax/linear form while preserving
quality. ReHyAt~\cite{ghafoorian2026rehyat} blends the two as well, but
computes the softmax component as a chunk-wise recurrence while using
linear attention for farther-away chunks, allowing constant runtime
memory.
A second, larger line keeps the exact softmax but evaluates it only over
a sparse subset of token pairs, exploiting the observation that video
attention maps are highly sparse and structured. VSA~\cite{zhang2025vsa}
makes this sparsity \emph{trainable}: a coarse stage pools tokens into
tiles to locate critical regions, and a fine stage computes token-level
attention only within them, as a single differentiable kernel usable in
both training and inference. Radial~Attention~\cite{li2025radial} instead
fixes a static $\mathcal{O}(n\log n)$ mask motivated by
``spatio-temporal energy decay'', shrinking each token's attention window
as temporal distance grows. 
% Sparse VideoGen2 (SVG2)~\cite{yang2025svg2} is training-free and improves critical-token selection by semantic-aware permutation, clustering and reordering tokens via k-means so that attention is computed over compact, semantically coherent blocks. 
Jenga~\cite{zhang2025jenga} is likewise training-free,
combining dynamic attention carving with space-filling curve-based token flattening at inference time. We compare
with these methods as our primary efficiency baselines, but note that \name 
differs from all of these in two ways. Unlike the
linear/hybrid surrogates, it retains a genuine softmax throughout; and
unlike the sparse-mask methods, its reduced communication pattern is
\emph{fixed} and structured,
% (global mixing across $\mathcal{O}(\sqrt{n})$ windows composed with local mixing withinthem), 
which lowers the complexity to $\mathcal{O}(n\sqrt{n})$
without the complex data-dependent token-selection steps.

\subsection{Distilling Video Generators}
\label{sec:relwork:distillation}
We find that for \name{}, a simple Flow-Matching~\cite{lipman2022flow}
SFT is \emph{not} enough to recover the
original's quality after swapping the softmax for \name-Attention; some
form of teacher distillation is required. For T2V generation,
such distillation is currently done in two complementary ways.

\noindent\textbf{Model (capability) distillation} matches what happens
\emph{inside} the teacher, not just what comes out of it. The idea is to
pull the student's intermediate features towards those of a stronger
network: REPA~\cite{yu2024repa} aligns a diffusion transformer's hidden
states with a self-supervised encoder like DINO, REPA-E~\cite{leng2025repae}
carries this through the VAE end-to-end, and \citet{wang2025repaworks}
study when such alignment helps and when to switch it off. All of this
leans on a now-familiar observation --- that diffusion backbones already
learn good, transferable features on their
own~\cite{xiang2023ddae}. Neodragon~\cite{karnewar2025neodragon} takes
the same idea to video, matching a teacher's activations to shrink a
model down for on-device generation.

\noindent\textbf{Step distillation} instead aims to reduce the NFEs needed at inference. 
Progressive
schemes halve the sampler's step count~\cite{salimans2022progressive,meng2023guided};
consistency-based methods map any trajectory point to its
endpoint~\cite{song2023consistency,luo2023lcm,kim2023ctm,lu2024scm};
adversarial schemes add a GAN objective for one-to-four-step
generation~\cite{sauer2023add,lin2025apt}; distribution-matching methods
minimize a reverse-KL between student and teacher via a score
difference~\cite{yin2024dmd,yin2024improved}, extended to video by
CausVid~\cite{yin2024causvid}, Self~Forcing~\cite{huang2025selfforcing},
and the Motion Consistency Model~\cite{zhai2024mcm}; and recent
flow-map methods learn the two-time map of the flow
directly~\cite{geng2025meanflow,sabour2025ayf,boffi2024flowmap}.

Rather than inventing yet another distillation method for our proposed \name-Attention, we simply adopt DMD2~\cite{yin2024improved}, which strikes a
good balance between the two objectives above: as a
distribution-matching method it allows us to recover the capability lost to
\name-Attention, while also reduces the step-count, allowing us to generate videos with far fewer NFEs.

% =====================================================================
%  SQuAD: Sub-Quadratic Attention Distillation for Efficient Video
%  Generation --- Method section.
%
%  This file is \input-able into a main paper. It assumes the packages
%  and macros defined in macros.tex are loaded (see main.tex for a
%  standalone compile).
% =====================================================================

\section{Method}

We first begin with full softmax Self-Attention, and fix the notation we build
on (\cref{sec:prelim-attn}) and then define the \name{}-Attention
operator (\cref{sec:squad-attn}). The broader discussion on latent video diffusion setup, and the details of how we train, are deferred to %\suppref{sec:supp_latent_video_diff} and % \suppref{sec:supp_training} 
sections A and B
of supplementary respectively. We recommend 
reading \suppref{sec:supp_latent_video_diff} prior to continuing.

% ---------------------------------------------------------------------

\subsection{Preliminaries: Full softmax Self-Attention}
\label{sec:prelim-attn}

% This is the method figure - part 2,  
% !!placing here for better readability!!
\begin{figure*}[t]
\centering
\includegraphics[width=\textwidth, trim=0 40 0 40, clip]{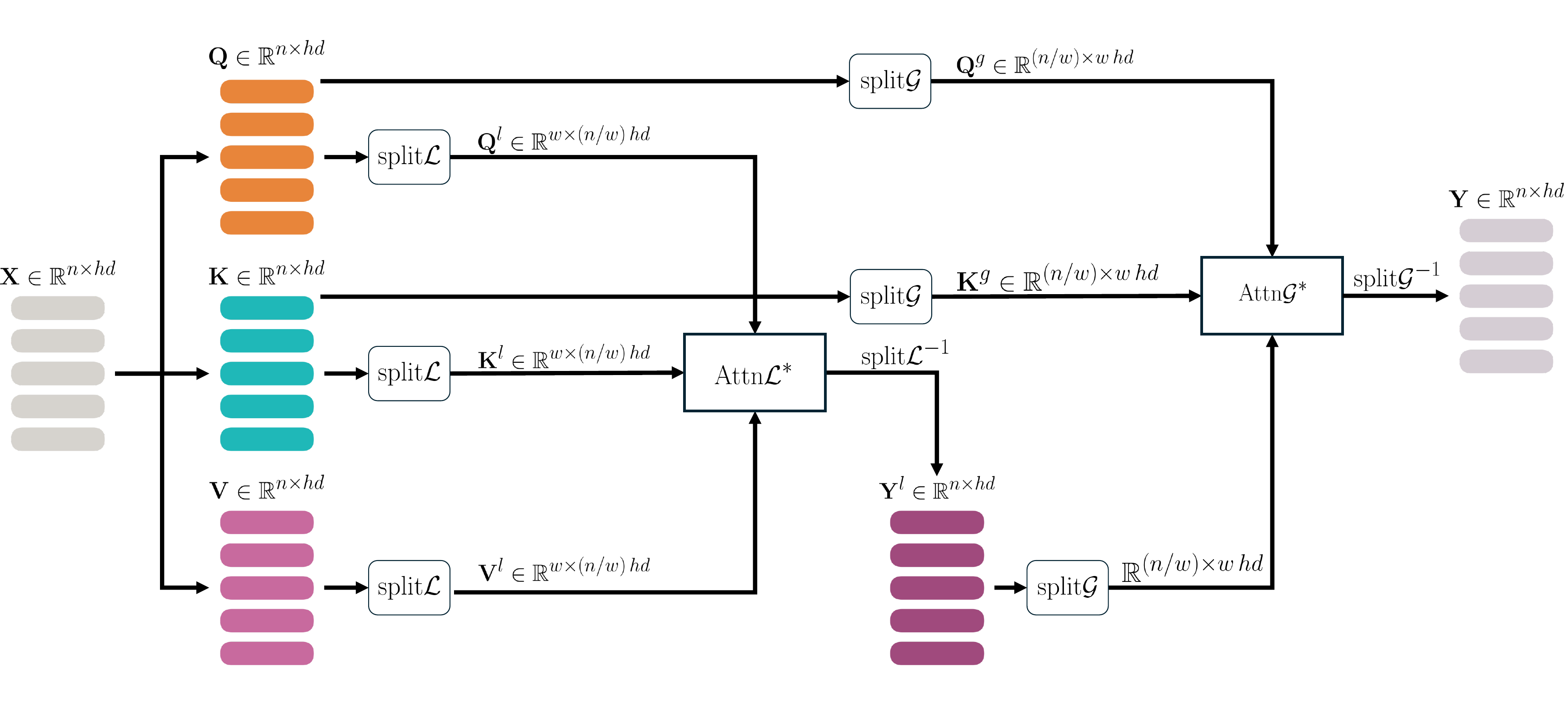} 
\caption{\textbf{Illustration of the \name Attention operation.}  The starred operators $\AttnL^{*}$ and $\AttnG^{*}$
apply the attention operations per-head and concatenate the outputs.}
\label{fig:method_part_2}
\end{figure*}

The DiT $\mathcal{D}_\theta$ is a composition of $L$ identical blocks
(with $L = 30$ for Wan~2.2 5B). The latent $z_\sigma$ is first flattened
and patchified into a sequence of $n$ token features, which the blocks
then refine in tandem:
\begin{equation}
  \begin{aligned}
    \bX^{(0)} &= \mathrm{embed}(z_\sigma), \\[2pt]
    \bX^{(\ell)} &= \mathcal{B}^{(\ell)}_\theta\!\big(\bX^{(\ell-1)}, \sigma, \mathbf{p}\big),
    \quad \ell = 1,\dots,L,
  \end{aligned}
  \label{eq:dit-blocks}
\end{equation}
where $\sigma$ is the current noise-level and $\mathbf{p}$ is the text prompt.
The flow-velocity is read out from $\bX^{(L)}$ by a final projection. Each
block $\mathcal{B}^{(\ell)}_\theta$ applies three residual sub-layers ---
a self-attention over the $n$ video tokens, a cross-attention to the text
prompt $\mathbf{p}$, and a token-wise feed-forward network --- each
modulated by the timestep $\sigma$ (omitted in following for brevity):
\begin{equation}
  \begin{aligned}
    \bX &\leftarrow \bX + \mathrm{SelfAttn}(\bX), \\
    \bX &\leftarrow \bX + \mathrm{CrossAttn}(\bX, \mathbf{p}), \\
    \bX &\leftarrow \bX + \mathrm{FFN}(\bX).
  \end{aligned}
  \label{eq:block}
\end{equation}
\name{} modifies only the self-attention while the cross-attention and the
feed-forward network are left untouched. In what follows we therefore
drop the block index and write $\bX \in \Real^{n \times hd}$ for the
token features entering a single block's self-attention.
This so-called Self-Attention operation that forms the backbone of essentially every
modern Transformer, diffusion or otherwise, is remarkably simple. A
sequence of $n$ input tokens is first
projected into three sequences queries, keys, and values as
\begin{equation}
  \bQ = \bX \bW_Q,\quad
  \bK = \bX \bW_K,\quad
  \bV = \bX \bW_V,
\end{equation}
with learned projection matrices
$\bW_{\{Q,K,V\}} \in \Real^{hd \times hd}$. Here $n$ is the number of
tokens; for video latents it is the size of a spatio-temporal volume,
$n = T \times H \times W$, as described earlier. Each of $\bQ$, $\bK$,
and $\bV$ is then split along its feature dimension into $h$ heads of
dimension $d$, giving per-head sequences
$\bQ_i, \bK_i, \bV_i \in \Real^{n \times d}$ for $i \in \{1,\dots,h\}$. A
rotary position embedding (RoPE)~\cite{su2021roformer} is applied to the
queries and keys only,
\begin{equation}
  \bQ_i \leftarrow \mathrm{RoPE}(\bQ_i), \qquad
  \bK_i \leftarrow \mathrm{RoPE}(\bK_i),
\end{equation}
while the values $\bV_i$ are left unchanged. The full softmax
self-attention for the $i$-th head is then
\begin{equation}
  \bY_i \;=\; \Attn(\bQ_i,\bK_i,\bV_i)
  \;=\;
  \softmax\!\left(\frac{\bQ_i \bK_i^{\top}}{\sqrt{d}}\right)\bV_i
  \label{eq:attn}
\end{equation}
Such that, $\softmax$ is a row-wise operator and
$\bY_i \in \Real^{n \times d}$. The per-head outputs are concatenated
back along the feature dimension,
\begin{equation}
  \bY \;=\; \mathrm{concat}(\bY_1,\dots,\bY_h) \;\in\; \Real^{n \times hd},
\end{equation}
yielding an output sequence $\bY$ with the same shape as the input
$\bX$. The $n \times n$ score matrix $\bQ_i\bK_i^{\top}$ inside the
softmax is the source of the quadratic cost: forming and applying it
takes $\bigO(n^2)$ work per head.

The Cross-Attention sub-layer is mechanically identical, except that the
keys and values are projected from the text-prompt embedding $\mathbf{p}$
rather than from $\bX$. Its cost is therefore cheap: the prompt length
($512$ tokens for Wan 2.2 5B) is typically far smaller than the number of visual
tokens $n$.

% ---------------------------------------------------------------------

\subsection{\name}
\label{sec:squad-attn}

Our proposal is a simple change. Where a standard DiT block
computes $\bY = \SelfAttn(\bX)$ via the full softmax $\Attn$ of
\cref{eq:attn}, we compute $\bY \;=\; \SQuadAttn(\bX)$. leaving every other 
part of the DiT blocks as it is.

\paragraph{A composition of two attentions.}
$\SQuadAttn$ is not a new kind of attention. It is the composition of two
ordinary softmax attentions, a \emph{local} pass $\AttnL$ and a
\emph{global} pass $\AttnG$. Writing the composition as a nesting makes
the data flow explicit,
\begin{equation}
  \bY_i
  \;=\;
  \underbrace{\AttnG\Big(
    \bQ_i,\; \bK_i,\;
    \underbrace{\AttnL\big(\bQ_i, \bK_i, \bV_i\big)}_{\mathclap{\bY^l_i}}
  \Big)}_{\mathclap{\bY^g_i}} .
  \label{eq:squad-compose}
\end{equation}
\Cref{fig:method_part1} gives the conceptual picture: the local pass mixes
tokens inside a window, and the global pass then mixes across windows.
Note that the two passes are not fused, and neither are their outputs pooled or summed or aggregated. Instead
the \emph{value stream acts as a shared scratch} between them. As
\cref{eq:squad-compose} shows, the local output enters the global pass
in the value slot. The global mixing therefore operates on tokens that
have already been locally mixed, and a full receptive field is recovered
even though neither pass alone is dense (see supplementary \suppref{sec:supp_receptive_field}).

\paragraph{Construction of the local and global token views.}
Both $\AttnL$ and $\AttnG$ are formally the $\Attn$ of \cref{eq:attn}, while they differ only
in how the queries and keys are laid out. Crucially, this re-viewing
happens \emph{before} the split into heads. Given a window size $w$, we
first rearrange the projected $\bQ, \bK, \bV \in \Real^{n \times hd}$
into a local and a global view,
\begin{equation}
  \begin{aligned}
    \{\bQ,\bK,\bV\}^{\loc}
      &= \splitL\big(\{\bQ,\bK,\bV\}\big)
       \in \Real^{w \times (n/w)\,hd},
    \\[2pt]
    \{\bQ,\bK\}^{\glob}
      &= \splitG\big(\{\bQ,\bK\}\big)
       \in \Real^{(n/w) \times w\,hd},
  \end{aligned}
  \label{eq:splits}
\end{equation}
and only then split into heads exactly as in \cref{sec:prelim-attn}. The
head dimension $d$ is unchanged; the number of heads absorbs the axis
that is no longer attended over. So the local view yields $(n/w)\,h$
heads over sequences of length $w$, and the global view yields $wh$
heads over sequences of length $(n/w)$. Both rearrangements are pure
re-indexings. They carry no parameters and no arithmetic, and each is
exactly invertible. We thus refer to their inverses $\splitL^{-1}$ and $\splitG^{-1}$
which simply revert the rearrangement of the tokens.

Concretely, for a video latent with  token grid of $T \times H \times W$ size and a window of
$w = w_t \times w_h \times w_w$ size, the two views are the \texttt{einops}
rearrangements
\begin{equation}
  \begin{aligned}
    \splitL:\;\;
      &\texttt{b (pt wt) (ph wh) (pw ww) h d}\\
      \rightarrow\;
      &\texttt{b (wt wh ww) (pt ph pw h) d},
    \\[6pt]
    \splitG:\;\;
      &\texttt{b (pt wt) (ph wh) (pw ww) h d}\\
      \rightarrow\;
      &\texttt{b (pt ph pw) (wt wh ww h) d},
  \end{aligned}
  \label{eq:einops}
\end{equation}
where \texttt{b} is the batch size, and $\texttt{pt} = T/w_t$, $\texttt{ph} = H/w_h$,
$\texttt{pw} = W/w_w$ count the windows along each axis. In both cases the second
axis of the result is the one attended over and the third collects the
remaining positions into the head count, leaving the head dimension
$d$ untouched. When the token extents are not divisible by the
window shape, our implementation pads the grid and masks the padded
positions in the attention. 

Given the definitions of the local and global split operations, both $\AttnL$ and $\AttnG$ have each the same three steps: split, attend, and unsplit,
\begin{equation}
  \begin{aligned}
    \AttnL(\cdot)
      &= \splitL^{-1}\Big(
           \Attn\big(\bQ_i^{\loc}, \bK_i^{\loc}, \bV_i^{\loc}\big)
         \Big),
    \\[4pt]
    \AttnG(\cdot)
      &= \splitG^{-1}\Big(
           \Attn\big(\bQ_i^{\glob}, \bK_i^{\glob}, \splitG(\bY_i^{\loc})\big)
         \Big).
  \end{aligned}
  \label{eq:attnLG}
\end{equation}
Both operators therefore consume and return tensors of the original shape
$\Real^{n \times hd}$. This is what makes them composable in
\cref{eq:squad-compose}. \Cref{fig:method_part_2} illustrates the flow of the operations 
mathematically defined in \cref{eq:attnLG}.

We analyze the resulting complexity in
\suppref{sec:supp_complexity}, and derive the optimal window size in \suppref{sec:supp_optimal_window} 
of the supplementary material. But even without the derivation, it can be intuitively observed 
that $w = \sqrt{n}$ is the
optimal choice for the window size balancing the number of tokens in the local and the global attention passes, ultimately giving us $\bigO(n\sqrt{n})$ Sub-Quadratic complexity.
Hence the name \name.

\paragraph{Distillation.}
Replacing $\SelfAttn$ with $\SQuadAttn$ changes the function each block
computes, so the pretrained weights no longer fit. Rather than train a
\name model from scratch, we demonstrate that a pretrained quadratic DiT
can be fitted into our modified one by distillation, in two stages. We
first perform a simple Flow-Matching SFT, which
re-seats the network under the new attention. We then apply DMD2 step
distillation, which recovers the teacher's generation quality and reduces
the sampling NFEs at the same time. We defer the details of both stages
to \suppref{sec:supp_training} of the supplementary.

\section{Experiments}

\definecolor{ourgreen}{RGB}{224,255,224}   % light green highlight for the DMD row
\definecolor{ourorange}{RGB}{252,230,217}  % light tint of #E76319
\definecolor{ourteal}{RGB}{216,235,238}    % light tint of #287380
\definecolor{ourgreencheck}{RGB}{80,190,120}  % green tick
\newcommand{\cmark}{\textcolor{ourgreencheck}{\ding{52}}}

% --- Table 1: full VBench (place at top, spans both columns) -------------------
\begin{table*}[t]
  \centering
  \caption{VBench evaluation comparing our \name to the original baseline, DMD distilled original, and various trained as well as training-free efficient attention methods. We show selected representative dimensions here due to space constraints.}
  \label{tab:vbench-full}
  \footnotesize
  \setlength{\tabcolsep}{3.45pt}
  \begin{tabularx}{\linewidth}{l ccc *{7}{c} *{6}{c}}
    \toprule
    \multirow{2}{*}{Method}
      & \multicolumn{3}{c}{VBench}
      & \multicolumn{7}{c}{Quality Dimensions}
      & \multicolumn{6}{c}{Semantic Dimensions} \\
    \cmidrule(lr){2-4} \cmidrule(lr){5-11} \cmidrule(lr){12-17}
      & \rotatebox{90}{Tot.}
      & \rotatebox{90}{Qual.}
      & \rotatebox{90}{Sem.}
      & \rotatebox{90}{Subj.}
      & \rotatebox{90}{Backg.}
      & \rotatebox{90}{T.Flick.}
      & \rotatebox{90}{Motion}
      & \rotatebox{90}{Dyn.}
      & \rotatebox{90}{Aesth.}
      & \rotatebox{90}{Imag.}
      & \rotatebox{90}{Obj.}
      & \rotatebox{90}{M-Obj.}
      & \rotatebox{90}{Act.}
      & \rotatebox{90}{Color}
      & \rotatebox{90}{Spat.}
      & \rotatebox{90}{Scene} \\
    \midrule
    \multicolumn{17}{c}{\cellcolor{black!8}\textbf{Base DiT: Wan 2.2 5B} \quad \textbf{Generation resolution: } \texttt{81x704x1280} \quad \textbf{Token length: } $n=\text{18480}$} \\
    \midrule
    \rowcolor{ourteal}
    Original & 83.08 & 83.98 & 79.48
      & 92.56 & 96.19 & 99.49 & 98.11 & 65.28 & 66.33 & 67.75
      & 89.08 & 71.02 & 97.20 & 83.68 & 81.83 & 54.10 \\
    DMD & 82.54 & 83.36 & 79.28
      & 94.70 & 94.69 & 96.73 & 96.89 & 75.83 & 66.94 & 69.00
      & 92.85 & 75.37 & 98.00 & 83.50 & 79.18 & 52.82 \\
    VSA & 84.14 & 84.93 & 81.00
      & 95.02 & 95.66 & 98.36 & 98.34 & 66.94 & 68.62 & 70.91
      & 94.53 & 81.14 & 98.80 & 84.77 & 85.71 & 53.65 \\
    Jenga & & & & & & & & & & & & & & & & \\
    \quad -- Training free & 83.08 & 84.14 & 78.82
      & 91.88 & 94.52 & 97.47 & 97.14 & 87.22 & 66.73 & 69.29
      & 90.49 & 71.57 & 98.20 & 84.93 & 79.06 & 51.12 \\
    \quad -- With training & 84.36 & 85.07 & 81.54
      & 94.44 & 95.84 & 99.01 & 97.72 & 71.94 & 68.60 & 70.18
      & 94.43 & 79.56 & 97.60 & 83.41 & 87.87 & 54.99 \\
    Radial Attention & 84.56 & 85.46 & 80.96
      & 94.44 & 95.85 & 98.77 & 97.08 & 84.17 & 68.03 & 69.98
      & 92.53 & 76.11 & 98.40 & 88.90 & 83.10 & 53.91 \\
    Attention Surgery & 83.39 & 84.55 & 78.74
      & 94.15 & 95.25 & 98.38 & 97.39 & 75.00 & 68.01 & 69.79
      & 92.04 & 72.56 & 97.60 & 83.12 & 82.50 & 51.21 \\
    ReHyAt & & & & & & & & & & & & & & & & \\
    \quad -- 20 Blocks & 83.70 & 84.46 & 80.66
      & 96.31 & 96.52 & 98.08 & 97.87 & 61.94 & 69.05 & 69.63
      & 94.87 & 81.31 & 97.40 & 84.69 & 82.10 & 53.94 \\
    \quad -- 30 Blocks & 83.22 & 83.62 & 81.61
      & 97.06 & 97.22 & 98.66 & 98.18 & 42.22 & 68.79 & 69.84
      & 95.73 & 84.45 & 97.00 & 83.80 & 84.90 & 55.99 \\
    \textbf{\name (Ours)} & & & & & & & & & & & & & & & & \\
    \quad -- 20 Blocks & 82.90 & 83.45 & 80.73
      & 95.12 & 95.64 & 97.31 & 97.27 & 65.28 & 68.21 & 68.93
      & 94.95 & 82.26 & 98.20 & 82.89 & 85.06 & 53.71 \\
    \rowcolor{ourorange}
    \quad -- 30 Blocks & 83.20 & 83.78 & 80.88
      & 95.77 & 96.41 & 97.79 & 98.16 & 57.22 & 68.70 & 68.47
      & 95.87 & 83.23 & 97.40 & 82.98 & 84.35 & 54.43 \\
    \midrule
    \multicolumn{17}{c}{\cellcolor{black!8}\textbf{Base DiT: Wan 2.1 1.3B} \quad \textbf{Generation resolution: } \texttt{81x480x832}\quad \textbf{Token length: } $n=\text{32760}$ }   \\
    \midrule
    Original & 83.26 & 84.25 & 79.30
      & 93.05 & 96.11 & 99.02 & 97.92 & 71.94 & 67.74 & 66.27
      & 90.41 & 74.39 & 96.80 & 86.35 & 77.87 & 52.71 \\
    DMD & 83.04 & 85.07 & 74.94
      & 93.94 & 94.78 & 97.29 & 97.77 & 85.28 & 68.51 & 70.06
      & 86.41 & 69.70 & 97.20 & 80.88 & 68.85 & 48.20 \\
    \textbf{\name (Ours)} & & & & & & & & & & & & & & & & \\
    \quad -- 20 Blocks & 82.74 & 84.43 & 76.00
      & 93.09 & 95.03 & 97.64 & 97.58 & 85.56 & 66.85 & 67.78
      & 89.21 & 75.09 & 97.20 & 82.45 & 70.44 & 44.30 \\
    \quad -- 30 Blocks & 82.70 & 84.27 & 76.44
      & 93.21 & 95.32 & 98.39 & 97.77 & 82.50 & 65.44 & 66.50
      & 91.61 & 75.00 & 95.20 & 85.76 & 70.48 & 45.62 \\
    \bottomrule
  \end{tabularx}
\end{table*}

\subsection{Experimental Setup}
\paragraph{Models.} We conduct all our experiments on the Wan 2.2 5B as well as the older Wan 2.1 1.3B \cite{wan2025}.

\paragraph{Metrics.} Our setup is designed to evaluate the efficacy of our proposed method over two axes: preservation of the generation quality, and improvement in the efficiency. We evaluate our experiments towards the first axis with the VBench \cite{huang2024vbench} suite of metrics as well as human user preference study, and along the second using the typical efficiency metrics such as TFLOPs, GPU latency, and NFEs. 

\paragraph{Human preference study.} To compare two methods, we randomly sampled long gpt-enhanced prompts from VBench and showed the users the prompts with a varying randomly ordered left/right videos and asked them to pick their preferred video or mark the comparison as a tie. This process resulted in 1,179 paired comparisons, from 24 distinct humans on 648 distinct prompts and 1,036 distinct videos, some with the same prompt but varying random seeds.

\paragraph{Datasets.} For our Stage-1 SFT we use the videos from the VIPE 1M dataset \cite{huang2025vipe}, and the descriptive captions for those generated by us using the Qwen3-8B-Instruct \cite{yang2025qwen3} model. We will release these captions for further research. For the Stage-2 DMD2 step distillation, we only use the captions, since this version of DMD bootstraps the student and does not require video data.

\subsection{Main Experiments}

We first present our evaluation of our proposed method \name in comparison to the original baseline models Wan 2.2 5B as well as the older Wan 2.1 1.3B since they operate on different resolutions natively, thus producing token lengths of different sizes. Specifically the token length for the 5B model at 704p resolution is $n=\text{18480}$ while for the 1.3B model at 480p resolution is $n=\text{32760}$. We experiment on  these two settings to demonstrate the generalization and robustness of our proposed change, while present a broader comparison against various efficient attention methods on the 5B model.

% \paragraph{Generation quality.}
As presented in \cref{tab:vbench-full} On Wan 2.2 5B, \name{} attains a VBench Total score of
$83.20$, effectively matching the full softmax Self-Attention of the original DiT
$83.08$. Crucially, it does so while introducing \textbf{zero additional
parameters} and with a simple implementation which is hardware device friendly by design,
not requiring any specialized GPU fused kernels or on-device optimizations. 
The competing efficient-attention method, Radial Attention \cite{li2025radial}, that posts higher VBech total (84.56) than ours
proposes an $\mathcal{O}(n\mathrm{log}(n))$ method, yet, as demonstrated in the \cref{tab:efficiency}, we surpass
their latency as well as TFLOPs without requiring specialized GPU kernels in the implementation. Lastly, as presented in
\cref{tab:user-study}, 31\% of the users couldn't distinguish between our and their generations while 35\% preferred ours. This strengthens the fact that our proposed method holds up to the compared efficient attention methods when it comes to video generation quality.

\Cref{tab:efficiency} shows that \name{} decisively improves efficiency from the original baseline.
On Wan 2.2 5B it reduces the per-block
latency from $62.01$ to $19.04$\,ms ($3.3\times$ faster), and compute from $9.681$
to $5.548$ TFLOPs ($1.7\times$ fewer), all with zero added parameters, unlike
VSA \cite{zhang2025vsa}, Attention Surgery \cite{ghafoorian2025attentionsurgery}, and ReHyAt \cite{ghafoorian2026rehyat}, which each carry $72$--$283$M extra weights. We note that the 
added learnable parameters is presented here merely as a representation of their method's complex nature. 
Our gains grow with model scale and sequence length.
Although we couldn't perform distillation experiments, we still run the efficiency evaluations 
on the larger Wan 2.1 14B backbone, and find that \name{} reduces compute from $41.959$ to $20.223$ TFLOPs and latency from
$300.29$ to $59.24$\,ms (ref \cref{tab:efficiency}). This is the
expected signature of trading an $\mathcal{O}(n^2)$ operation for an
$\mathcal{O}(n\sqrt{n})$ one: the deeper the token grid, the larger the payoff.
Lastly, beyond the primary Wan 2.2 5B model, we apply the identical \name distillation
to Wan 2.1 1.3B to test that the method is not tied to a single
backbone or single token-length setting. The efficiency results (\Cref{tab:efficiency}) transfer directly;
the corresponding Wan 2.1 quality evaluations are reported in
\Cref{tab:vbench-full}.

% % --- Table 3: User study between our model and the originals ------
% \begin{table}[t]
%   \centering
%   \caption{User study. Base model for all is Wan 2.2 5B.}
%   \label{tab:user-study}
%   \footnotesize
%   \setlength{\tabcolsep}{6pt}
%   \renewcommand{\arraystretch}{1.25}
%   \begin{tabularx}{\columnwidth}{l c c c}
%     \toprule
%     \multirow{2}{*}{Baseline Method}
%       & \multicolumn{3}{c}{Human Preference \%} \\
%     \cmidrule(lr){2-4}
%       & \name & No pref. & Baseline \\
%     \midrule
%     Original 100 NFEs            & 41\% & 33\% & 26\% \\
%     DMD 6 NFEs                   & 33\% & 37\% & 30\% \\
%     DMD Rad. Attn. 6 NFEs        & 35\% & 31\% & 34\% \\
%     \bottomrule
%   \end{tabularx}
% \end{table}

% % --- Table 4: SFT/DMD ablation --- 30 Blocks (main paper) --------------------
% \begin{table}[t]
%   \centering
%   \caption{Ablation of SFT and DMD for \name on 30 Blocks.}
%   \label{tab:sft-dmd-ablation-30}
%   \footnotesize
%   \renewcommand{\arraystretch}{1.25}
%     \begin{tabular*}{\columnwidth}{@{\extracolsep{\fill}} c c c c c c}
%     \toprule
%     \multirow{2}{*}{SFT}
%       & \multirow{2}{*}{DMD}
%       & \multirow{2}{*}{NFE}
%       & \multicolumn{3}{c}{VBench} \\
%     \cmidrule(lr){4-6}
%       & & & Tot. & Qual. & Sem. \\
%     \midrule
%     \cmark &        & 100 & 73.03 & 75.66 & 62.48 \\
%            & \cmark & 6   & 80.91 & 82.23 & 75.65 \\
%     \cmark & \cmark & 6   & \textbf{82.99} & \textbf{83.69} & \textbf{80.19} \\
%     \bottomrule
%     \end{tabular*}
% \end{table}
\begin{table}[t] \centering \begin{minipage}[t]{0.48\columnwidth} \centering \caption{User study. Base model for all is Wan 2.2 5B.} \label{tab:user-study} \footnotesize \setlength{\tabcolsep}{4pt} \renewcommand{\arraystretch}{1.15} \begin{tabular}{lccc} \toprule \multirow{2}{*}{Baseline Method} & \multicolumn{3}{c}{Human Preference \%} \\ \cmidrule(lr){2-4} & \name & No pref. & Baseline \\ \midrule Original 100 NFEs & 41\% & 33\% & 26\% \\ DMD 6 NFEs & 33\% & 37\% & 30\% \\ DMD Rad. Attn. 6 NFEs & 35\% & 31\% & 34\% \\ \bottomrule \end{tabular} \end{minipage} \hfill \begin{minipage}[t]{0.48\columnwidth} \centering \caption{Ablation of SFT and DMD for \name on 30 Blocks.} \label{tab:sft-dmd-ablation-30} \footnotesize \renewcommand{\arraystretch}{1.15} \begin{tabular}{cccccc} \toprule \multirow{2}{*}{SFT} & \multirow{2}{*}{DMD} & \multirow{2}{*}{NFE} & \multicolumn{3}{c}{VBench} \\ \cmidrule(lr){4-6} & & & Tot. & Qual. & Sem. \\ \midrule \cmark & & 100 & 73.03 & 75.66 & 62.48 \\ & \cmark & 6 & 80.91 & 82.23 & 75.65 \\ \cmark & \cmark & 6 & \textbf{82.99} & \textbf{83.69} & \textbf{80.19} \\ \bottomrule \end{tabular} \end{minipage} \end{table}

\subsection{Hardware Compilation and Performance}
\label{sec:supp_latency-protocol}

The complexity analysis of \cref{sec:supp_complexity} establishes that \name reduces
the attention cost from $\mathcal{O}(n^2)$ to $\mathcal{O}(n\sqrt{n})$, and the
Block-level TFLOPs counts as well as the Block-level Latencies of \cref{tab:efficiency} 
confirm that this translates into a real reduction in arithmetic work. 
Neither quantity, however, is what a practitioner actually waits for. 
Asymptotic complexity ignores constants, memory traffic, and kernel efficiency; 
FLOP counts credit every multiply-add equally, even though a 
large fused GEMM and a bandwidth-bound softmax over an
$n \times n$ score matrix reach wildly different fractions of peak throughput.
An attention operator can therefore look cheap on paper and still fail to
deliver, either because the work it removes was never the bottleneck or because
the reshaping it introduces costs more than the multiplications it saves. The
only way to settle the question is to measure wall-clock time on the device. We
report that measurement here, in both of the execution modes a user is likely to
deploy: PyTorch \emph{eager} mode and a \texttt{torch.compile} \emph{compiled}
mode in \cref{tab:supp_latency}.

% % --- Table X: end-to-end DiT latency (eager vs compiled) ----------------------
% \begin{table}[t]
%   \centering
%   \caption{End-to-end latency of a single DiT forward pass, measured in PyTorch
%            eager mode and under \texttt{torch.compile} (\texttt{max-autotune}).
%            All rows are the Wan 2.2 5B backbone at \texttt{81x704x1280}.
%            }
%   \label{tab:supp_latency}
%   \footnotesize
%   \renewcommand{\arraystretch}{1.2}
%   \begin{tabularx}{\columnwidth}{X c c c}
%     \toprule
%     \multirow{2}{*}{\textbf{Method}}
%       & \multirow{2}{*}{\textbf{NFE}}
%       & \multicolumn{2}{c}{\textbf{End-to-End DiT Latency}} \\
%     \cmidrule(lr){3-4}
%       & & \textbf{Eager mode} & \textbf{Compiled mode} \\
%     \midrule
%     \rowcolor{ourteal}
%     Original DiT              & 100 & 870ms    & 667ms    \\
%     VSA                       & 6   & 724ms    & 434ms    \\
%     Jenga                     & 6   & 680ms    & 427ms    \\
%     Radial Attention          & 6   & 765ms    & 619ms    \\
%     % SVG2                    & 6   & \dots    & \dots    \\
%     Attention Surgery         & 6   & 1006ms   & 537ms    \\  % - $R8$
%     ReHyAt                    & 6   & 1757ms   & 544ms    \\  % - $T_c\!=\!3;T_o\!=\!1$
%     \rowcolor{ourorange}
%     \textbf{\name (Ours)}     & 6   & \textbf{520ms} & \textbf{314ms} \\
%     \bottomrule
%   \end{tabularx}
% \end{table}

\paragraph{What we time.}
We time a \emph{single forward pass of the DiT} --- the quantity that the
sampler invokes once per function evaluation --- rather than the whole
generation pipeline. Text encoding and VAE decoding are excluded: they are
shared verbatim by every method we compare and would otherwise dilute the
effect under study. Timing uses pairs of
\texttt{torch.cuda.Event(enable\_timing=True)} objects registered as
forward-pre and forward hooks on the transformer module. Because the hooks sit
on the outermost module wrapper and only \emph{record} events, they measure the
compiled region without entering it, and so remain valid under
\texttt{fullgraph=True} compilation; the same instrumentation is used unchanged
in both modes. Each measured interval is bracketed on the CUDA stream and read
back only after a \texttt{torch.cuda.synchronize()}, so the reported numbers are
true device latencies and contain no host-side asynchrony artifacts. Classifier-free
guidance is disabled ($\omega = 0$; recall from \cref{sec:supp_training} that CFG
is distilled \emph{into} the student), so each denoising step issues exactly one
transformer forward and our NFE count equals the number of timed forwards.

\begin{table}[t] \centering \newsavebox{\effbox} \newsavebox{\latbox} \savebox{\effbox}{\scriptsize \setlength{\tabcolsep}{4pt} \renewcommand{\arraystretch}{1.2} \begin{tabular}{l c c c c} \toprule \textbf{Method} & \textbf{NFE} & \textbf{\#Par.} & \textbf{Lat.} & \textbf{TFLOPs} \\ \midrule \multicolumn{5}{c}{\cellcolor{black!8}\textbf{Base DiT: Wan 2.2 5B}\; (\texttt{81x704x1280})} \\ \midrule \rowcolor{ourteal} Original DiT & 100 & 0 & 62.01ms & 9.7 \\ VSA & 6 & 283M & 24.13ms & 6.2 \\ Jenga & 6 & 0 & 22.63ms & 6.2 \\ Radial Attention & 6 & 0 & 25.50ms & 7.0 \\ Attention Surgery & 6 & 72M & 33.53ms & 6.8 \\ ReHyAt & 6 & 72M & 58.56ms & 6.5 \\ \rowcolor{ourorange} \textbf{\name (Ours)} & 6 & 0 & 19.04ms & 5.5 \\ \midrule \multicolumn{5}{c}{\cellcolor{black!8}\textbf{Other variants}\; (\texttt{81x480x832})} \\ \midrule Wan 2.1 1.3B Orig. & 100 & 0 & 88.36ms & 9.4 \\ \quad -- \textbf{\name (Ours)} & 6 & 0 & 15.22ms & 2.9 \\ \addlinespace[2pt] Wan 2.1 14B Orig. & 100 & 0 & 300.29ms & 41.9 \\ \quad -- \textbf{\name (Ours)} & 6 & 0 & 59.24ms & 20.2 \\ \bottomrule \end{tabular}} \savebox{\latbox}{\scriptsize \setlength{\tabcolsep}{4pt} \renewcommand{\arraystretch}{1.2} \begin{tabular}{l c c c} \toprule \multirow{2}{*}{\textbf{Method}} & \multirow{2}{*}{\textbf{NFE}} & \multicolumn{2}{c}{\textbf{End-to-End DiT Latency}} \\ \cmidrule(lr){3-4} & & \textbf{Eager} & \textbf{Compiled} \\ \midrule \rowcolor{ourteal} Original DiT & 100 & 870ms & 667ms \\ VSA & 6 & 724ms & 434ms \\ Jenga & 6 & 680ms & 427ms \\ Radial Attention & 6 & 765ms & 619ms \\ Attention Surgery & 6 & 1006ms & 537ms \\ ReHyAt & 6 & 1757ms & 544ms \\ \rowcolor{ourorange} \textbf{\name (Ours)} & 6 & \textbf{520ms} & \textbf{314ms} \\ \bottomrule \end{tabular}} \begin{minipage}[t]{\wd\effbox} \caption{Efficiency comparisons} \label{tab:efficiency} \usebox{\effbox} \end{minipage} \hspace{2em} \begin{minipage}[t]{\wd\latbox} \caption{End-to-end latency of a single DiT forward pass, measured in PyTorch eager mode and under \texttt{torch.compile} (\texttt{max-autotune}). All rows are the Wan 2.2 5B backbone at \texttt{81x704x1280}.} \label{tab:supp_latency} \usebox{\latbox} \end{minipage} \end{table}

\paragraph{Protocol and statistics.}
We profile a fixed list of $30$ prompts spanning varied scenes, subjects, and
camera motion. The first $5$ prompts are executed but discarded: they populate
the allocator caches, trigger cuDNN/cuBLAS autotuning, and---in compiled
mode---absorb the one-off graph capture and \texttt{max-autotune} kernel search,
none of which reflect steady-state cost. The remaining $25$ prompts are
\emph{counted}, and with $\text{NFE} = 6$ this yields
$25 \times 6 = 150$ timed forward passes per mode, which we pool before
aggregating. We report the mean over this pooled sample.
% , and additionally track
% the median, the $90$th percentile, and a two-sided $5\%$ trimmed standard
% deviation (dropping $\lceil 0.05N \rceil$ samples from each tail of the sorted
% list) so that the spread is not dominated by isolated scheduling outliers. 
% The
The RNG is re-seeded before every prompt, so a given prompt sees identical noise in
the eager and compiled passes and the two modes are compared on exactly the same
work. Every method in \cref{tab:supp_latency} is measured with this identical
harness, on the same machine, in the same session-level environment.

\subsection{Ablation Experiments}

We now finally detail the process of experimentation through which we found the best configuration 
of our proposed \name method reported in \cref{tab:vbench-full}. In principle, all design choices 
following the $\mathcal{O}(n\sqrt{n})$ framework work reasonably, but since we are not training from 
scratch and distilling a pretrained model, some choices work better. Since it made more conceptual sense to us, 
we actually started the experiments with the global\,$\to$\,local order of the proposed attentions. 

\paragraph{Both training stages are necessary. For now.}
Thus, on the global\,$\to$\,local order of application, \Cref{tab:sft-dmd-ablation-30} disentangles 
Stage-1 SFT from Stage-2 DMD2. 
It is clear that when all $30$ blocks of the model are replaced with \name-Attention: SFT alone collapses (VBench Tot. $73.03$), and DMD2 alone reaches only $80.91$, but the two stages \emph{together} recover the full
$82.99$ VBench Tot. 

\paragraph{Window characterization.}
Then the next question is what shape or characterization of the windows should we use. 
\Cref{tab:window-characterization} studies how the choice of window
geometry, at a roughly fixed token budget near the sub-quadratic target
$\lceil\sqrt{n}\,\rceil = \text{136}$, for $n=\text{18480}$ affects quality, contrasting aspect-ratio
preserving, spatial-only, temporal-only, and isotropic windows. We find that the temporal windows are 
the easiest to train and produce the best quality videos. This could be attributed to the asymetric 
nature of the DiT patchification (1x temporal while 2x spatial), but further analysis is required to reason properly.

\paragraph{The need for and the order of the two  passes.}
Finally, we attempt to experimentally answer the question: what order should the two passes run in.
\Cref{tab:local-global-ablation-30} shows that, used in
isolation, the local-only and global-only passes are the cheapest
% ($168\times$ attention-FLOPs reduction for global-only) 
but don't work properly
% and at $30$ blocks they collapse outright (Total $62.62$ and $62.63$
% respectively) ---
as the restricted receptive field limits their generation quality strongly. 
Composing the two restores full quality, and a uniform local\,$\to$\,global pass is the best with VBench total of $83.20$ 
% ahead of global\,$\to$\,local VBench of $82.99$. 
We also experimented with blockwise alternating order of the two passes, and found it to perform worse than the uniform local\,$\to$\,global pass.

% --- Table 5: local/global ordering, 30 blocks (spans both columns) -------------
\begin{table*}[t]
  \centering
  \caption{Ablation of local/global attention and their ordering on \name applied to all 30 Blocks. Attention latencies are in ms and $\times$ columns are improvements relative to the Original Attention.}
  \label{tab:local-global-ablation-30}
  \footnotesize
  \setlength{\tabcolsep}{7pt}
  \renewcommand{\arraystretch}{1.25}
  \begin{tabularx}{\textwidth}{X c c c c c c c c c c c c}
    \toprule
    \multirow{2}{*}{Method}
      & \multirow{2}{*}{Local}
      & \multirow{2}{*}{Global}
      & \multirow{2}{*}{Order}
      & \multirow{2}{*}{\shortstack{Block\\TFLOPs}}
      & \multirow{2}{*}{\shortstack{Block\\Lat.}}
      & \multirow{2}{*}{\shortstack{Attn.\\TFLOPs}}
      & \multirow{2}{*}{\shortstack{Attn.\\Lat.}}
      & \multirow{2}{*}{\shortstack{TFLOPs\\$\times$}}
      & \multirow{2}{*}{\shortstack{Lat.\\$\times$}}
      & \multicolumn{3}{c}{VBench} \\
    \cmidrule(lr){11-13}
      & & & & & & & & & & Tot. & Qual. & Sem. \\
    \midrule
    \rowcolor{ourteal}
    Original                     & --     & --     & --        & 9.680 & 61.80 & 4.205 & 47.10 & 1.0   & 1.0   & 83.08 & 83.98 & 79.48 \\
    \midrule
    \name                        & \cmark &        & --        & 5.523 & 19.46 & 0.038 & 3.61  & 110.7 & 13.0  & 62.62 & 73.65 & 18.50 \\
                                 &        & \cmark & --        & 5.509 & 17.86 & 0.025 & 2.88  & 168.2 & 16.4  & 62.63 & 71.48 & 27.24 \\
    \rowcolor{ourorange}
    \quad (Ours)                 & \cmark & \cmark & L$\to$G   & 5.548 & 19.55 & 0.063 & 4.27  & 66.7  & 11.0  & 83.20 & 83.78                              & 80.88 \\
                                 & \cmark & \cmark & G$\to$L   & 5.548 & 19.93 & 0.063 & 4.25  & 66.7  & 11.1  & 82.99 & 83.69 & 80.19 \\
                                 & \cmark & \cmark & alternate & --    & --    & --    & --    & --    & --    & 82.87 & 83.57 & 80.06 \\
    \bottomrule
  \end{tabularx}
\end{table*}

% --- Table 6: window characterization -----------------------------------------
\begin{table}[t]
  \centering
  \caption{Window characterization ablation. Here $N = 18480$ tokens, so the
           window target is $\lceil\sqrt{N}\,\rceil = 136$; ``Target diff.''\ is
           each window's \#Tokens minus this target.}
  \label{tab:window-characterization}
  \footnotesize
  \setlength{\tabcolsep}{6pt}
  \renewcommand{\arraystretch}{1.25}
  \begin{tabularx}{\columnwidth}{l c c c c}
    \toprule
    Window shape
      & Dims
      & \#Tok.
      & \shortstack{diff.} % Target\\
      & \shortstack{VBench Tot.} \\ %\\Tot.
    \midrule
    AR-preserving      & $4\times5\times7$   & 140 & $+4$  & 81.99 \\
    Spat.              & $1\times9\times16$  & 144 & $+8$  & 82.31 \\
    Spat. isotropic    & $1\times12\times12$ & 144 & $+8$  & 81.55 \\
    Temp.     & $21\times2\times4$  & 168 & $+32$ & \textbf{82.99} \\
    Temp. isotropic    & $21\times3\times3$  & 189 & $+53$ & 81.80 \\
    Isotropic (small)  & $5\times5\times5$   & 125 & $-11$ & 81.55 \\
    Isotropic (large)  & $6\times6\times6$   & 216 & $+80$ & 81.69 \\
    \bottomrule
  \end{tabularx}
\end{table}

\section{Conclusion}
We introduced \name{}, a \namef{} framework that makes the attention in a
pretrained video DiT sub-quadratic \emph{without} abandoning the
softmax that gives it its expressivity. Beyond the concrete efficiency gains, 
% --- matching the quadratic teacher on VBench ($83.20$ vs.\ $83.08$) while cutting
% per-step attention FLOPs by ${\sim}67\times$ and attention latency by
% ${\sim}11\times$, and generating in $6$ NFEs instead of $100$ --- 
our broader aim
is to shine light on a simple hypothesis: the full $\mathcal{O}(n^2)$ complexity of
token-to-token communication may not be necessary. Cheaper
\emph{factorizations that keep the softmax intact}, rather than replace it with
linear or low-rank surrogates, are an under-explored middle ground. We hope
\name{} encourages this direction not only as a distillation target for
existing models, but also as a candidate attention mechanism for pretraining large
models from scratch.

\paragraph{Limitations and future work.}
% Our results should be read as an early demonstration, and several directions
% follow naturally. 
First, our distillation strategy couples the \name{} attention with
\emph{step} distillation (DMD2); a version that stays purely within
Flow-Matching, without step reduction, is the most immediate next step and would
isolate the attention change from the sampling speedup. Second, 
our proposed \name applies two passes of reduced complexity attention, but we hypothesize that
% \name{}'s
% $\mathcal{O}(n\sqrt{n})$ complexity, while sub-quadratic, is still above the
% $\mathcal{O}(n\log n)$ of divide-and-conquer schemes; 
composing \emph{more} than
the two passes could, in principle, trade a little more depth for a
cheaper and more expressive attention operator. Finally, \name{} is not
specific to video: applying this simple idea to other domains and modalities
remains an interesting and insightful avenue.

% For an arXiv upload, replace the next line with \input{main.bbl} and
% ship the generated .bbl, or ship the .bbl alongside and let arXiv
% pick it up. arXiv does not re-run BibTeX against bibs/aaai2027.bib.

% ---- Supplement ------------------------------------------------------
% z__supplementary.tex issues its own \appendix and its own A-prefixed
% figure/table/equation numbering, and starts a new page through the
% \twocolumn shim above.
\ifsupp
  % =====================================================================
%  Supplementary material -- CONTENT ONLY.
%  \input from main.tex inside the \ifsupp branch. Do not add a
%  \documentclass / \begin{document} here.
% =====================================================================

% Start the supplement on a fresh page with a FULL-WIDTH title block.
% \twocolumn[...] breaks the page and typesets its argument across both
% columns -- the same trick \maketitle uses -- then resumes two-column
% body text. NOTE: this is only ever executed in the combined arXiv
% build, never in the AAAI submission build, so it cannot violate the
% submission style rules.
% (Inside \twocolumn[...] we use \par + \vspace rather than \\[Xem],
%  because \\ is fragile in this argument.)
\twocolumn[%
  \begin{center}
    {\LARGE\bfseries \name{}: \namef{} for Efficient Video Generation\par}
    \vspace{1.0em}
    {\Large\bfseries Supplementary Material\par}
  \end{center}
  \vspace{1.5em}
]

% Appendix-style numbering: sections become A, B, ...; figures, tables
% and equations get an "A" prefix so they can never collide with the
% main paper's numbering.
\appendix
\setcounter{section}{0}
\setcounter{figure}{0}
\setcounter{table}{0}
\setcounter{equation}{0}
\renewcommand{\thefigure}{A\arabic{figure}}
\renewcommand{\thetable}{A\arabic{table}}
\renewcommand{\theequation}{A\arabic{equation}}

% ---------------------------------------------------------------------
\section{Preliminaries: Latent Video Diffusion Transformers}
\label{sec:supp_latent_video_diff}
Modern text-to-video generators are, almost universally, latent video
Diffusion Transformers (DiTs) trained in two stages~\citesupp{rombach2022ldm}:
a variational autoencoder is first learned as a \emph{continuous
tokenizer}, and a DiT is then trained to generate in that continuous
tokenized latent space with a flow-matching objective.

\paragraph{Stage 1 --- continuous tokenizer.}
A causal 3D variational autoencoder is trained to compress and
reconstruct videos. An encoder $\mathcal{E}_\psi$ maps an RGB clip
$Z \in \Real^{3 \times F \times H_p \times W_p}$ (with $F$ frames of
spatial size $H_p \times W_p$) to the parameters of a Gaussian latent
posterior,
\begin{equation}
  (\bm{\mu}, \bm{s}) \;=\; \mathcal{E}_\psi(Z), \qquad
  z \;=\; \bm{\mu} + \bm{s} \odot \bm{\epsilon}',\;\;
  \bm{\epsilon}' \sim \normal(\zero, \bI),
  \label{eq:supp_reparam}
\end{equation}
where the reparameterized latent
$z \in \Real^{c \times T \times H \times W}$ is far smaller than $Z$, and
a decoder $\mathcal{G}_\psi$ reconstructs the video, $\hat{Z} =
\mathcal{G}_\psi(z)$. The pair is trained end-to-end with a
reconstruction loss and a KL term that pulls the posterior towards a
unit Gaussian,
\begin{equation}
  \loss_{\mathrm{VAE}}
  \;=\;
  \underbrace{\big\| Z - \mathcal{G}_\psi(z) \big\|}_{\text{reconstruction}}
  \;+\;
  \beta\,
  \underbrace{D_{\mathrm{KL}}\!\big(\normal(\bm{\mu}, \bm{s}^2) \,\big\|\, \normal(\zero, \bI)\big)}_{\text{latent regularization}}.
  \label{eq:supp_vae}
\end{equation}
In practice $\loss_{\mathrm{VAE}}$ is often augmented with perceptual and
adversarial losses to train more robust tokenizers. For Wan~2.2 5B the
VAE compresses by ${\sim}4\times$ temporally and $16\times$ spatially
into $c = 48$ latent channels. Once trained, it is frozen and serves as
the tokenizer for Stage~2: the latent grid $T \times H \times W$ is
flattened into a sequence of $n = T H W$ tokens that the DiT operates on.

\paragraph{Stage 2 --- flow-matching DiT.}
A DiT $\mathcal{D}_\theta$ is trained over the frozen latents with the
rectified-flow (flow-matching) objective~\citesupp{lipman2022flow}. Taking a
clean latent $z_0$ from the Stage-1 encoder and Gaussian noise
$\bm{\epsilon} \sim \normal(\zero, \bI)$, the noisy latent at level
$\sigma \in [0,1]$ is the straight-line interpolant
\begin{equation}
  z_\sigma \;=\; (1-\sigma)\,z_0 + \sigma\,\bm{\epsilon},
  \label{eq:supp_interp}
\end{equation}
and $\mathcal{D}_\theta$ is trained to predict the constant velocity
$\bm{\epsilon} - z_0$ of this path which points from the data
towards the unit Gaussian by minimizing
\begin{equation}
  \loss_{\mathrm{FM}}
  \;=\;
  \E_{z_0,\,\bm{\epsilon},\,\sigma}
  \big\|\, \mathcal{D}_\theta(z_\sigma, \sigma, \mathbf{p}) - (\bm{\epsilon} - z_0) \,\big\|_2^2,
  \label{eq:supp_fm}
\end{equation}
where $\mathbf{p}$ is the text-prompt embedding. In practice $z_0$ is a
reparameterized sample $z$ (\cref{eq:supp_reparam}) for robustness, though
using the posterior mean $\bm{\mu}$ instead also works comparably.

\paragraph{Sampling.}
At inference, generation runs the flow in reverse. Starting from pure
noise $z_1 \sim \normal(\zero, \bI)$ and conditioned on a prompt
$\mathbf{p}$, we integrate the learned velocity field from $\sigma = 1$
to $\sigma = 0$ with iterative Euler steps,
\begin{equation}
  z_{\sigma - \Delta\sigma}
  \;=\;
  z_\sigma - \Delta\sigma\, \mathcal{D}_\theta(z_\sigma, \sigma, \mathbf{p}),
  \label{eq:supp_euler}
\end{equation}
over a schedule of Neural Functional Evaluations (NFEs) to obtain the
clean latent $z_0$. The decoder $\mathcal{G}_\psi$ finally maps it back
to an RGB video, $\hat{Z} = \mathcal{G}_\psi(z_0)$. We note that several
further ingrained details such as patchification of the latent grid,
and classifier-free guidance (which doubles the number of Euler-step
NFEs) are standard de-facto practice; we omit them here for lucidity,
but note that they are present and duly handled in all our experiments.

% ---------------------------------------------------------------------
\section{Method: Two-stage distillation}
\label{sec:supp_training}

Replacing full attention with \name{} changes the function the network
computes, so the pretrained weights no longer fit. We recover ---and
then accelerate--- the model in two stages. Stage~1 re-fits the \name{}
network to the teacher's behaviour with the original flow-matching
objective. Stage~2 distills the many-step model into a few-step
generator with distribution-matching distillation. Throughout, the
frozen pretrained Wan~2.2 model serves as the teacher.

\paragraph{Stage 1: flow-matching supervised fine-tuning.}
We instantiate the \name{} transformer $\mathcal{D}_\theta$ from the
pretrained weights and fine-tune it with the same rectified-flow loss
used to train the backbone. For a clean latent $z_0$, noise
$\bm{\epsilon}\sim\normal(\zero,\bI)$, and a sampled noise level
$\sigma$, we form the interpolant $z_\sigma$ of \cref{eq:supp_interp} and
regress the velocity:
\begin{equation}
  \loss_{\mathrm{SFT}}(\theta)
  \;=\;
  \E_{z_0,\,\bm{\epsilon},\,\sigma}
  \Big\|\,
    \mathcal{D}_\theta(z_\sigma, \sigma, \mathbf{p})
    \;-\;
    \big(\bm{\epsilon} - z_0\big)
  \,\Big\|_2^{2}.
  \label{eq:supp_sft}
\end{equation}
The noise level $\sigma$ is drawn from a logit-normal schedule. This
stage adapts the query/key/value and output projections --- and the rest
of the network --- to the new attention pattern, restoring generation
quality under a standard 50-step sampler. We train Stage~1 for $8$k
iterations.

\paragraph{Stage 2: distribution-matching distillation.}
Stage~1 yields a competitive but still multi-step model. To make it
fast we distill it into a few-step generator with DMD2. Three networks
are involved: the few-step \emph{student} generator $G_\theta$
(initialized from the Stage~1 \name{} model), a frozen \emph{teacher}
$\mathcal{D}^{\mathrm{tea}}_\phi$ (the pretrained model, giving the real
score), and a trainable \emph{critic} $\mathcal{D}^{\mathrm{fake}}_\psi$
(a copy of the \name{} model that tracks the student's own distribution).

The student maps noise to a clean latent in a few steps; denote a
generated sample $\hat{z}_0$.
Following the flow parametrization, from a prediction at level $\sigma$
we read off the implied clean latent
$\hat{z}_0 = z_\sigma - \sigma\, \mathcal{D}_\theta(z_\sigma,\sigma,\mathbf{p})$.
DMD matches the distribution of $\hat{z}_0$ to the teacher's by
pushing the student down the gradient of the KL divergence between the
two, which reduces to the difference of two scores evaluated on a
re-noised sample
$\hat{z}_\tau = (1-\tau)\hat{z}_0 + \tau\,\bm{\epsilon}'$
at a randomly sampled level $\tau$:
\begin{equation}
  \nabla_\theta \mathrm{KL}
  \;\propto\;
  \E\Big[\,
    \big(
      s_{\mathrm{fake}}(\hat{z}_\tau,\tau)
      \;-\;
      s_{\mathrm{real}}(\hat{z}_\tau,\tau)
    \big)\,
    \frac{\partial \hat{z}_0}{\partial \theta}
  \Big].
  \label{eq:supp_dmd-grad}
\end{equation}
Here $s_{\mathrm{fake}}$ is the critic's velocity/score on the student
distribution and $s_{\mathrm{real}}$ is the teacher's. The teacher score
uses classifier-free guidance,
\begin{equation}
  s_{\mathrm{real}}
  \;=\;
  \mathcal{D}^{\mathrm{tea}}_\phi(\hat{z}_\tau,\tau,\varnothing)
  \;+\;
  \omega\,\big(
    \mathcal{D}^{\mathrm{tea}}_\phi(\hat{z}_\tau,\tau,\mathbf{p})
    - \mathcal{D}^{\mathrm{tea}}_\phi(\hat{z}_\tau,\tau,\varnothing)
  \big),
  \label{eq:supp_cfg}
\end{equation}
with guidance scale $\omega$ and $\varnothing$ the null prompt. In
practice we normalize the distribution-matching gradient by its
magnitude and inject it through a stop-gradient surrogate loss, so that
$-\nabla_\theta \loss_{\mathrm{DMD}}$ equals the gradient in
\cref{eq:supp_dmd-grad}.

The critic is what makes \cref{eq:supp_dmd-grad} usable: it must estimate
the score of the \emph{current} student distribution, which drifts as
$\theta$ updates. We therefore train $\mathcal{D}^{\mathrm{fake}}_\psi$
online, by flow-matching on the student's own samples,
\begin{equation}
  \loss_{\mathrm{critic}}(\psi)
  \;=\;
  \E\Big\|\,
    \mathcal{D}^{\mathrm{fake}}_\psi(\hat{z}_\tau, \tau, \mathbf{p})
    \;-\;
    \big(\bm{\epsilon}' - \hat{z}_0\big)
  \,\Big\|_2^{2}.
  \label{eq:supp_critic}
\end{equation}
Generator and critic are optimized on a two-timescale schedule: the
critic is updated several times per generator update so that its score
estimate stays accurate. We run Stage~2 for
$15$--$30$k iterations, distilling down to a $6$-step generator. Because
classifier-free guidance is itself distilled into the student, the
reported NFEs are literal forward passes of the network.

% ---------------------------------------------------------------------
\section{Related Work: Text-to-Video Generation}
\label{sec:supp_relwork}
Diffusion models have rapidly become the dominant paradigm for
text-to-video generation. The transition from U-Net backbones to the
Diffusion Transformer (DiT)~\citesupp{peebles2023dit} architecture, which
replaces convolutions with a scalable stack of self-attention blocks
operating on spatio-temporal latent tokens, has been central to the
recent leap in visual quality and temporal coherence. Building on this
method, a broad family of large-scale video generators has emerged,
including HunyuanVideo~\citesupp{kong2024hunyuanvideo},
Open-Sora~\citesupp{zheng2024opensora}, Pyramidal
Flow~\citesupp{jin2024pyramidal}, LTX-Video~\citesupp{hacohen2024ltxvideo},
Cosmos~\citesupp{nvidia2025cosmos}, Neodragon~\citesupp{karnewar2025neodragon},
and the Wan~2.1/2.2 suite~\citesupp{wan2025}. We conduct our experiments primarily on the Wan~2.2
model, chosen for its architectural relevance, simplicity, and, above
all, its popularity as an open video foundation model.

% ---------------------------------------------------------------------
\section{Full Receptive Field in a Single \name Layer}
\label{sec:supp_receptive_field}

In \cref{sec:squad-attn} we claimed that although neither the local pass
$\AttnL$ nor the global pass $\AttnG$ is dense on its own, their
composition recovers a full receptive field within a single layer. We
prove that here.

\paragraph{Indexing.}
It is convenient to name tokens by their position in the windowed view
rather than by their flat index. Let $p = n/w$ be the number of windows
and index every token by the pair $(c, j)$, where $c \in \{1,\dots,p\}$
is its window and $j \in \{1,\dots,w\}$ is its slot inside that window.
The two rearrangements of \cref{eq:splits} attend over exactly one of
these indices each: $\splitL$ attends over $j$ at fixed $c$, and
$\splitG$ attends over $c$ at fixed $j$. Writing the resulting softmax
weights for head $h$ as
\begin{equation}
  \beta^{\,c,h}_{j \to j'}
  \;=\;
  \softmax_{j'}\!\left(
    \tfrac{1}{\sqrt{d}}\,
    \langle \bQ_{c,j,h},\, \bK_{c,j',h}\rangle
  \right)
  \label{eq:supp_beta}
\end{equation}
for the local pass, and
\begin{equation}
  \alpha^{\,j,h}_{c \to c'}
  \;=\;
  \softmax_{c'}\!\left(
    \tfrac{1}{\sqrt{d}}\,
    \langle \bQ_{c,j,h},\, \bK_{c',j,h}\rangle
  \right)
  \label{eq:supp_alpha}
\end{equation}
for the global pass, the two passes read
\begin{equation}
  \begin{aligned}
    \big(\bY_l\big)_{c,j,h}
      &= \sum_{j'=1}^{w} \beta^{\,c,h}_{j \to j'}\;\bV_{c,j',h},
    \\[2pt]
    \big(\bY_g\big)_{c,j,h}
      &= \sum_{c'=1}^{p} \alpha^{\,j,h}_{c \to c'}\;\big(\bY_l\big)_{c',j,h}.
  \end{aligned}
  \label{eq:supp_passes}
\end{equation}
Each is a proper softmax attention: the weights are non-negative and
$\sum_{j'} \beta^{\,c,h}_{j \to j'} = \sum_{c'} \alpha^{\,j,h}_{c \to c'} = 1$.

\paragraph{Unrolling the composition.}
Substituting the first line of \cref{eq:supp_passes} into the second
gives the output of a single \name layer at token $(c,j)$,
\begin{equation}
  \begin{aligned}
    \big(\bY_g\big)_{c,j,h}
      &= \sum_{c'=1}^{p} \alpha^{\,j,h}_{c \to c'}
         \sum_{j'=1}^{w} \beta^{\,c',h}_{j \to j'}\;\bV_{c',j',h}
    \\[4pt]
      &= \sum_{c'=1}^{p} \sum_{j'=1}^{w}
         \underbrace{\alpha^{\,j,h}_{c \to c'}\,\beta^{\,c',h}_{j \to j'}}
                    {\textstyle \gamma^{\,h}_{(c,j) \to (c',j')}}
         \;\bV_{c',j',h}.
  \end{aligned}
  \label{eq:supp_unrolled}
\end{equation}
The double sum ranges over \emph{all} $p \cdot w = n$ tokens. So the
output at $(c,j)$ is a weighted average of every value in the sequence,
with effective weights
$\gamma^{\,h}_{(c,j)\to(c',j')} = \alpha^{\,j,h}_{c\to c'}\,\beta^{\,c',h}_{j\to j'}$.
These form a valid distribution as well, since
\begin{equation}
  \sum_{c'}\sum_{j'} \gamma^{\,h}_{(c,j)\to(c',j')}
  \;=\;
  \sum_{c'} \alpha^{\,j,h}_{c \to c'} \sum_{j'} \beta^{\,c',h}_{j \to j'}
  \;=\;
  1 .
\end{equation}

\paragraph{Reachability.}
Concretely, information travels from any source token $(c',j')$ to any
target token $(c,j)$ along a two-hop path. The local pass first carries
$(c',j')$ to $(c', j)$, moving within the source window to the slot $j$
that the target occupies. The global pass then carries $(c', j)$ to
$(c, j)$, moving across windows at that shared slot. Both hops exist for
every choice of indices, so every source reaches every target.

Two remarks follow. First, the effective weight
$\gamma^{\,h}_{(c,j)\to(c',j')}$ is a product of two softmax weights, so
it is strictly positive whenever both hops are, meaning the layer has no
structural zeros in its receptive field. Second, the reachability is
achieved in one layer, not accumulated over depth as in strictly local
attention such as sliding windows. \name is therefore a drop-in
replacement for full attention rather than a local approximation of it,
which is what allows a pretrained quadratic model to be fitted to it by
distillation alone.

\paragraph{On the ordering.}
The argument is symmetric in the two passes. Running global before local
gives
$\gamma^{\,h}_{(c,j)\to(c',j')} = \beta^{\,c,h}_{j\to j'}\,\alpha^{\,j',h}_{c\to c'}$,
which is again a full receptive field, differing only in which
intermediate token the two hops route through: $(c,j')$ instead of
$(c',j)$. The two orders are therefore equally dense but not identical
as functions, which is what motivates the ordering ablation of
\cref{tab:local-global-ablation-30}.

% ---------------------------------------------------------------------
\section{Complexity of \name Attention}
\label{sec:supp_complexity}

We derive here the cost of a single \name layer as a function of the
window size $w$, counting the work spent forming and applying the softmax
score matrices. Throughout, $n$ is the sequence length, $h$ the number of
heads, $d$ the head dimension, and $p = n/w$ the number of windows.

\paragraph{Cost of one softmax attention.}
For a batch of $B$ independent attentions, each over a sequence of length
$s$ with head dimension $d$, the two matrix products inside
\cref{eq:attn} cost
\begin{equation}
  \underbrace{B \cdot s^2 \cdot d}_{\bQ\bK^{\top}}
  \;+\;
  \underbrace{B \cdot s^2 \cdot d}_{\text{applying to }\bV}
  \;=\;
  \bigO\!\big(B\,s^2\,d\big).
  \label{eq:supp_attn-cost}
\end{equation}
We use this as the unit of accounting. The full softmax attention of
\cref{sec:prelim-attn} is the case $B = h$, $s = n$, giving the familiar
$\bigO(h\,n^2\,d)$.

\paragraph{Cost of the two passes.}
The rearrangements of \cref{eq:einops} determine $B$ and $s$ for each
pass. Recall that the attended axis becomes the sequence, and the
remaining positions are folded into the head count.

For the \emph{local} pass, the sequence is the within-window position, so
$s = w$, and the head count absorbs the $p$ windows, so $B = p\,h$.
Substituting into \cref{eq:supp_attn-cost},
\begin{equation}
  \mathcal{C}_{\mathcal{L}}
  \;=\;
  \underbrace{(p\,h)}_{\text{attentions}} \cdot
  \underbrace{w^2}_{\text{length}^2} \cdot
  \underbrace{d}_{\text{feat.}}
  \;=\;
  h\,d\;\frac{n}{w}\,w^2
  \;=\;
  h\,d\,n\,w .
  \label{eq:supp_local-cost}
\end{equation}

For the \emph{global} pass, the sequence is the window index, so
$s = p = n/w$, and the head count absorbs the $w$ within-window slots, so
$B = w\,h$. This gives
\begin{equation}
  \mathcal{C}_{\mathcal{G}}
  \;=\;
  \underbrace{(w\,h)}_{\text{attentions}} \cdot
  \underbrace{\left(\frac{n}{w}\right)^{\!2}}_{\text{length}^2} \cdot
  \underbrace{d}_{\text{feat.}}
  \;=\;
  h\,d\,\frac{n^2}{w} .
  \label{eq:supp_global-cost}
\end{equation}

\paragraph{Total.}
Since \name composes the two passes, their costs add, and the total
attention cost of a \name layer is
\begin{equation}
  \mathcal{C}(w)
  \;=\;
  \mathcal{C}_{\mathcal{L}} + \mathcal{C}_{\mathcal{G}}
  \;=\;
  h\,d\left(n\,w + \frac{n^2}{w}\right).
  \label{eq:supp_total-cost}
\end{equation}
The two terms pull in opposite directions. Enlarging the window makes the
local pass more expensive (longer sequences within a window) and the
global pass cheaper (fewer windows to attend across); shrinking it does
the reverse. Note also the two extremes: $w = n$ places every token in a
single window and recovers full quadratic attention through the local
pass, while $w = 1$ does the same through the global pass. The
interesting regime is in between.

% ---------------------------------------------------------------------
\section{The Optimal Window Size}
\label{sec:supp_optimal_window}

Because \cref{eq:supp_total-cost} is a sum of one term increasing in $w$
and one decreasing in $w$, it has an interior minimum. Treating $w$ as
continuous and differentiating,
\begin{equation}
  \frac{\mathrm{d}\mathcal{C}}{\mathrm{d}w}
  \;=\;
  h\,d\left(n - \frac{n^2}{w^2}\right),
  \label{eq:supp_derivative}
\end{equation}
and setting this to zero gives
\begin{equation}
  n \;=\; \frac{n^2}{w^2}
  \quad\Longrightarrow\quad
  w^2 \;=\; n
  \quad\Longrightarrow\quad
  w^\star \;=\; \sqrt{n}.
  \label{eq:supp_optimum}
\end{equation}
This is a minimum rather than a maximum, since
$\mathrm{d}^2\mathcal{C}/\mathrm{d}w^2 = 2\,h\,d\,n^2/w^3 > 0$ for all
$w > 0$. Equivalently, $w^\star$ is the point at which the two terms of
\cref{eq:supp_total-cost} balance: substituting $w = \sqrt{n}$ makes both
$n\,w$ and $n^2/w$ equal to $n^{3/2}$.

\paragraph{Resulting complexity.}
Evaluating \cref{eq:supp_total-cost} at the optimum,
\begin{equation}
  \mathcal{C}(w^\star)
  \;=\;
  h\,d\left(n\sqrt{n} + \frac{n^2}{\sqrt{n}}\right)
  \;=\;
  2\,h\,d\,n^{3/2}
  \;=\;
  \bigO\!\big(n\sqrt{n}\big).
  \label{eq:supp_optimal-cost}
\end{equation}
Setting the window size to $\sqrt{n}$ is therefore exactly the regime in
which \name attains sub-quadratic $\bigO(n\sqrt{n})$ complexity, at the
cost of only two softmax passes instead of one. This is the configuration
of our best model. In the video setting $w$ must additionally factor into
a valid window shape $w_t w_h w_w$, so we pick the admissible shape
closest to $\sqrt{n}$ while preserving the aspect ratio of the latent
volume; \cref{tab:local-global-ablation-30} and our window-shape ablation
study the effect of departing from this choice.

% --- Table: local/global ordering, 20 blocks (spans both columns) -------------
\begin{table*}[t]
  \centering
  \caption{Ablation of local/global attention and their ordering (\name 20 Blocks). Best results are in \textbf{bold}. Attention TFLOPs/Latency are residuals (attn1 minus its projection/norm submodules); latencies are in ms and $\times$ columns are improvements relative to the Original.}
  \label{tab:local-global-ablation-20}
  \footnotesize
  \setlength{\tabcolsep}{4pt}
  \renewcommand{\arraystretch}{1.25}
  \begin{tabularx}{\textwidth}{X c c c c c c c c c c c c}
    \toprule
    \multirow{2}{*}{Method}
      & \multirow{2}{*}{Local}
      & \multirow{2}{*}{Global}
      & \multirow{2}{*}{Order}
      & \multirow{2}{*}{\shortstack{Block\\TFLOPs}}
      & \multirow{2}{*}{\shortstack{Block\\Lat.}}
      & \multirow{2}{*}{\shortstack{Attn.\\TFLOPs}}
      & \multirow{2}{*}{\shortstack{Attn.\\Lat.}}
      & \multirow{2}{*}{\shortstack{TFLOPs\\$\times$}}
      & \multirow{2}{*}{\shortstack{Lat.\\$\times$}}
      & \multicolumn{3}{c}{VBench} \\
    \cmidrule(lr){11-13}
      & & & & & & & & & & Tot. & Qual. & Sem. \\
    \midrule
    \rowcolor{ourteal}
    Original                     & --     & --     & --        & 9.680 & 61.80 & 4.205 & 47.10 & 1.0   & 1.0   & 83.08 & 83.98 & 79.48 \\
    \midrule
    \name 20 Blocks              & \cmark &        & --        & 5.523 & 19.46 & 0.038 & 3.61  & 110.7 & 13.0  & 81.86 & 83.64 & 74.74 \\
                                 &        & \cmark & --        & 5.509 & 17.86 & 0.025 & 2.88  & 168.2 & 16.4  & 82.31 & 82.98 & 79.61 \\
    \rowcolor{ourorange}
                                 & \cmark & \cmark & L$\to$G   & 5.548 & 19.55 & 0.063 & 4.27  & 66.7  & 11.0  & 82.90 & 83.44 & 80.73 \\
                                 & \cmark & \cmark & G$\to$L   & 5.548 & 19.93 & 0.063 & 4.25  & 66.7  & 11.1  & 82.73 & 83.46 & 79.76 \\
                                 & \cmark & \cmark & alternate & --    & --    & --    & --    & --    & --    & 82.86 & 83.39 & 80.73 \\
    \bottomrule
  \end{tabularx}
\end{table*}

\paragraph{Hardware and software environment.}
All latencies are measured on a single \textbf{NVIDIA H100 80\,GB HBM3}
(compute capability $9.0$, $132$ streaming multiprocessors, $79.2$\,GiB of
usable HBM3), with one GPU visible to the process so that no measurement is
perturbed by co-tenancy. The software stack is PyTorch $2.9.1$ (CUDA build
$12.8$, cuDNN $9.10.2$) with \texttt{diffusers} $0.36.0$, and the caching
allocator is configured with
\texttt{expandable\_segments:True} to avoid fragmentation
stalls across the $30$-prompt sweep. Weights and all attention computation are
in \texttt{bfloat16}. The compiled mode wraps the transformer with
\texttt{torch.compile(dynamic=False, fullgraph=True, mode="max-autotune")};
static shapes are appropriate here because the latent grid is fixed by the
generation resolution. Measurements are taken at our headline generation setting
of $81 \times 704 \times 1280$ (i.e.\ $n = 18480$ latent tokens) with
$\text{NFE} = 6$, and peak memory is recorded alongside each run
($\approx 60$\,GB in both modes for \name).

\paragraph{Analyzing the results.}
\Cref{tab:supp_latency} reports the mean single-forward latency for every method
in both modes, and the two columns tell complementary stories. \name is the
fastest method in \emph{both}: at $520$\,ms eager and $314$\,ms compiled it is
$1.67\times$ and $2.12\times$ faster than the full-attention Original, and it
still leads the closest efficient baseline (Jenga, at $680$ and $427$\,ms) by
$1.31\times$ and $1.36\times$ respectively. That the advantage holds in both
columns is the point: a ranking established in one execution mode need not
survive the other, since compilation fuses pointwise work and removes Python
dispatch overhead \emph{unevenly} across methods.

The eager column is where this unevenness becomes diagnostic. Two of the
baselines are actually \emph{slower} than the $870$\,ms full-attention Original
that they set out to accelerate, despite reporting lower FLOP counts in
\cref{tab:efficiency}: Attention Surgery costs $1006$\,ms and ReHyAt $1757$\,ms.
Their speedups materialize only after compilation, at $537$ and $544$\,ms
respectively, which is to say that their theoretical savings are not realizable
by the stock PyTorch executor. They depend instead on specialized, hand-tuned GPU
kernels, such as custom sparse-attention or chunked-recurrence implementations,
to convert a FLOP reduction into a wall-clock reduction. This is a real
deployment cost. Such kernels must be written, maintained, and re-tuned for each
new accelerator generation, and until they exist for a given target the
advertised efficiency simply does not appear.

\name requires none of this. Its two passes are ordinary softmax attentions
separated by pure re-indexing (\cref{eq:squad-compose}), so the modified transformer is
handed \emph{unchanged} to the most vanilla compilation setting PyTorch offers, a
single \texttt{torch.compile} call, with no custom CUDA, no bespoke attention
kernel, and no operator registration. The $1.66\times$ that compilation adds on
top of the architectural gain therefore comes for free, from a compiler that
knows nothing about \name. We take this as the practical counterpart to the
method's conceptual simplicity: an efficiency gain that is available to anyone
who can call \texttt{torch.compile} is far more likely to be adopted than one
gated behind a kernel-engineering effort, and it transfers to new hardware the
moment the compiler does.

\section{FLOPs Scaling Analysis}
\label{sec:flops-scaling}

\begin{figure*}[t]
  \centering
  \begin{subfigure}[t]{0.33\textwidth}
    \includegraphics[width=\linewidth]{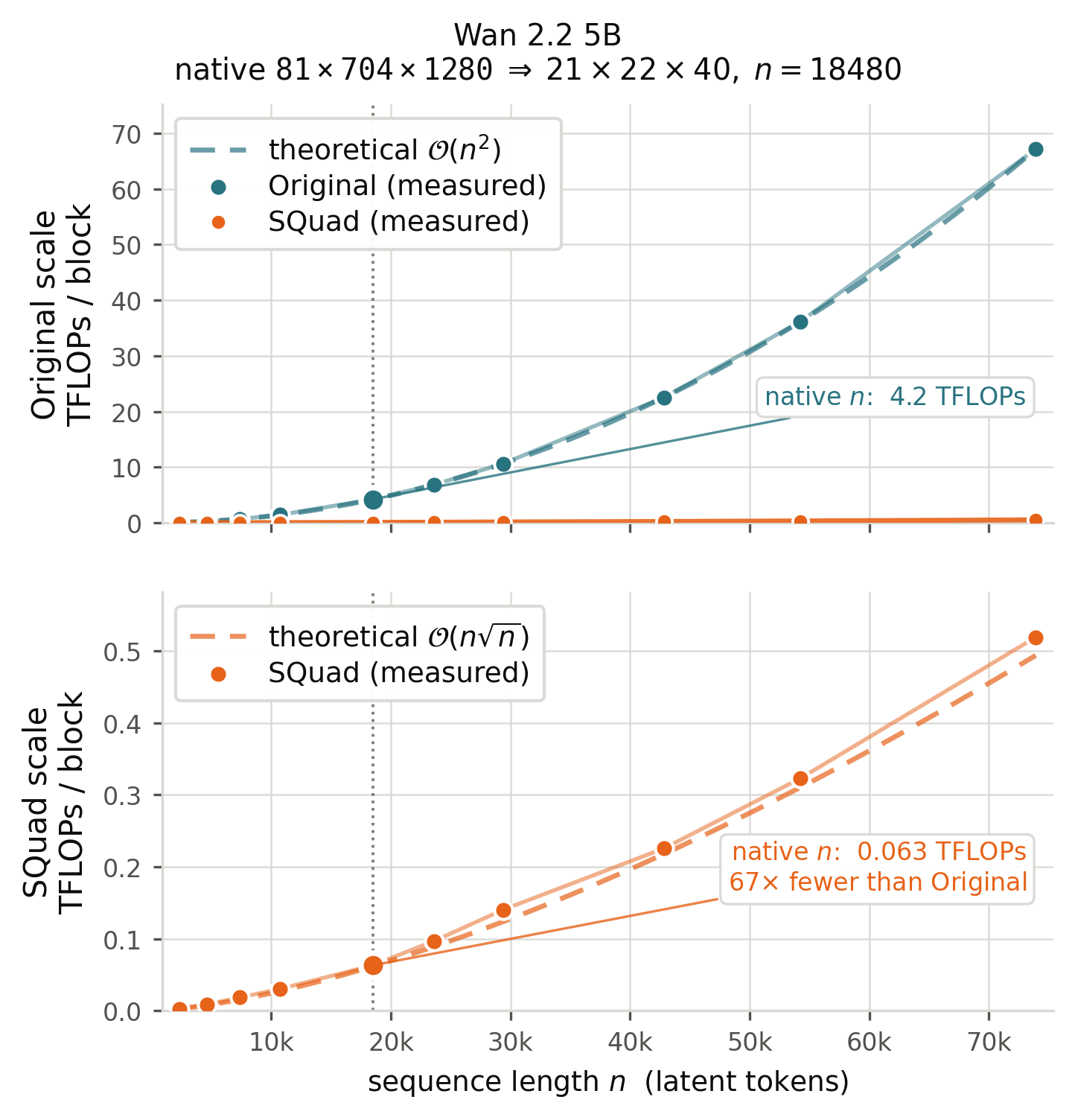}
    \caption{Wan 2.2 5B}
    \label{fig:flops-scaling-5b}
  \end{subfigure}
  \hfill
  \begin{subfigure}[t]{0.33\textwidth}
    \includegraphics[width=\linewidth]{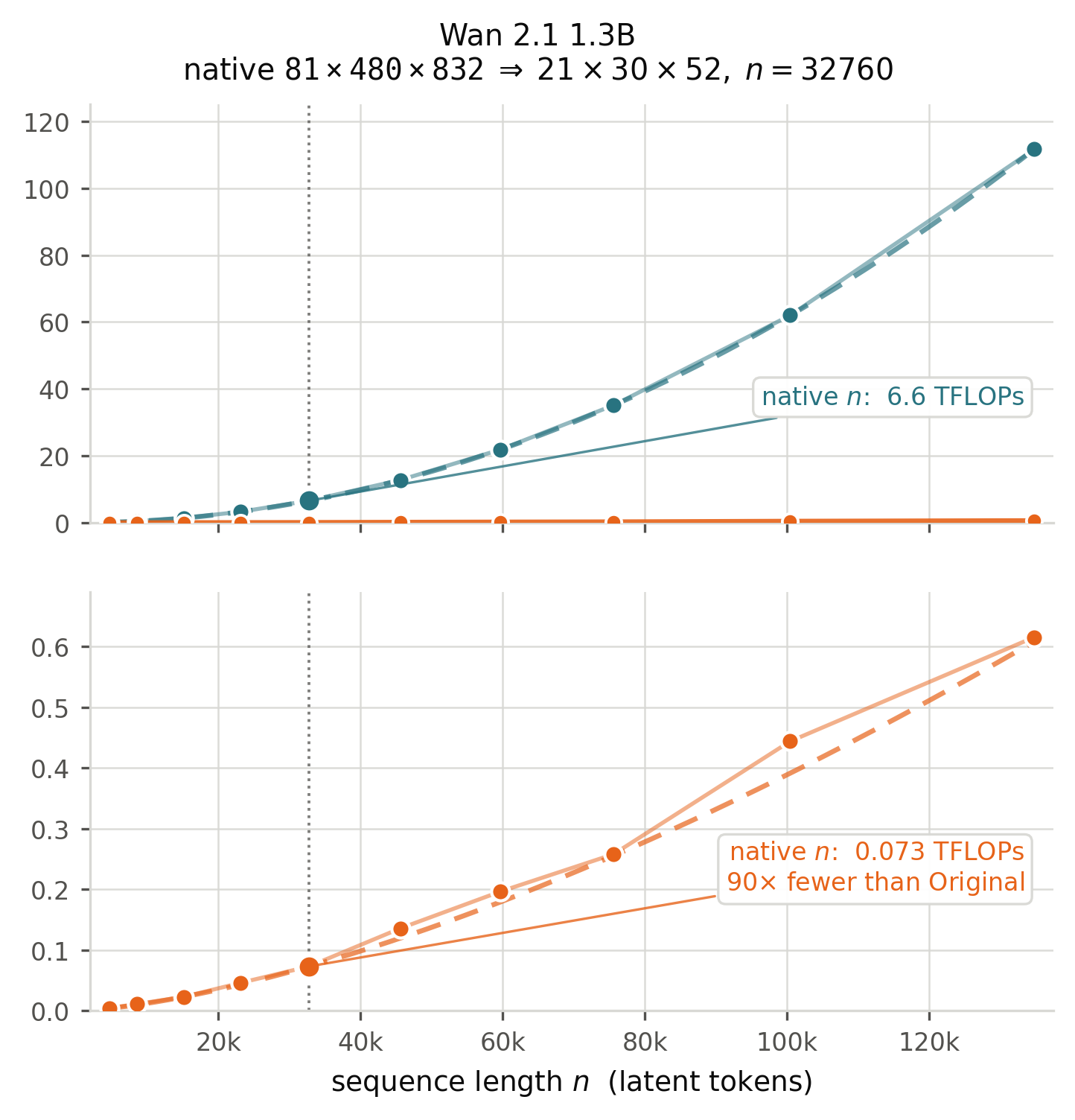}
    \caption{Wan 2.1 1.3B}
    \label{fig:flops-scaling-1p3b}
  \end{subfigure}
  \hfill
  \begin{subfigure}[t]{0.33\textwidth}
    \includegraphics[width=\linewidth]{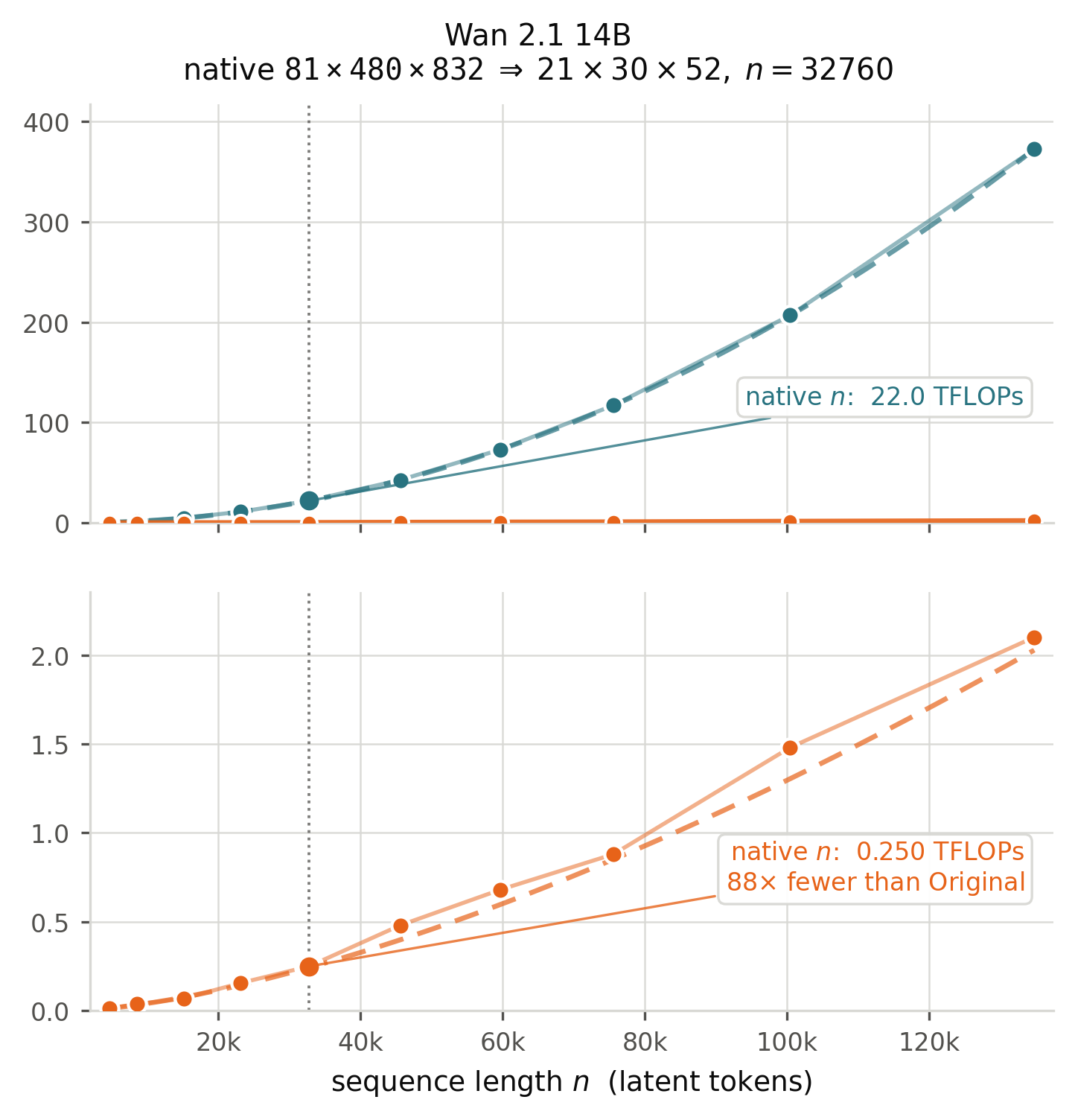}
    \caption{Wan 2.1 14B}
    \label{fig:flops-scaling-14b}
  \end{subfigure}
  \caption{Attention FLOPs per block vs.\ sequence length $n$, across three
    backbones. Top: full-softmax Original against its $\mathcal{O}(n^2)$
    reference, with \name overlaid. Bottom: \name on its own scale against the
    $\mathcal{O}(n\sqrt{n})$ reference. Markers are measured; dashed lines are
    the theoretical curves.}
  \label{fig:supp_flops-scaling}
\end{figure*}

The complexity result of \cref{sec:supp_complexity} is asymptotic, and the latency
study of \cref{sec:supp_latency-protocol} is a single operating point. This section
sits between the two: it measures the actual attention FLOPs of a single block,
as a function of sequence length $n$, across the full range of resolutions each
backbone could plausibly generate. The question it answers is whether the
$\mathcal{O}(n\sqrt{n})$ promise is realized in practice, or whether constants
erode it.

\paragraph{What is measured.}
For each backbone we build one real transformer block in each of its two forms,
Original and \name, and profile the Self-Attention cost with DeepSpeed's
FLOPs profiler. We sweep ten real video shapes per model at the native aspect ratio,
spanning roughly a $30\times$ range in $n$, and repeat the sweep for all three
backbones. Crucially, the window is re-solved at every resolution rather than held
fixed, since \name is $\mathcal{O}(n\sqrt{n})$ only when the window volume $m$
tracks $\sqrt{n}$. We keep the two constraints the method itself imposes: every
window spans the full temporal extent ($w_t = T$), so temporal mixing stays exact
and only the spatial extents grow with $n$; and $w_h : w_w$ follows the latent
grid's aspect ratio, so windows remain geometrically similar to the frame instead
of degenerating into strips. Reassuringly, this rule independently recovers the
trained $21\times2\times4$ window at both backbones' native resolutions, which
confirms that the sweep and the trained models describe one method.
\Cref{fig:supp_flops-scaling} plots the result; every marker is a genuine
measurement, and the dashed line through each provides the theoretical function
curve for reference.

\paragraph{The measured curves match the theory.}
The Original attention overlays its $\mathcal{O}(n^2)$ reference almost exactly
across the entire range, on all three backbones. This is a useful sanity check on
the measurement itself: the profiled quadratic cost is the quadratic cost the
analysis predicts, with no hidden constant drift. The \name curve, plotted on its
own scale in the lower panel of each figure, follows the $\mathcal{O}(n\sqrt{n})$
reference closely across the whole sweep and grows far more slowly than the
quadratic curve everywhere. Read on the shared Original scale (upper panels),
\name is a nearly flat line pinned to zero, which is the visual form of the
reduction: at the native operating point it costs $0.063$, $0.073$, and $0.250$
TFLOPs for the 5B, 1.3B, and 14B backbones, against $4.2$, $6.6$, and $22.0$
TFLOPs for full attention, i.e.\ $67\times$, $90\times$, and $88\times$ fewer.

\paragraph{Conformance to the theoretical curves is quantitative.}
Because both references are absolute rather than fitted, the agreement can be
stated as a ratio instead of read off a plot. Across all thirty measured points
the Original cost lands within $0.4\%$ of $4hdn^2$, so the quadratic baseline is
matched essentially exactly. The \name measurements sit within $21\%$ of the
idealized $\mathcal{O}(n\sqrt{n})$ optimum and $8\%$ above it on average, with no
systematic drift as $n$ grows: the largest and smallest sequence lengths conform
equally well, which is the signature of a genuinely sub-quadratic operator rather
than a quadratic one with a small constant. The residual gap is fully explained by
integrality, and not by any missing term in the analysis. The optimum $m^\star =
\sqrt{n}$ is generally irrational, whereas a realizable window must satisfy $m =
T \cdot w_h \cdot w_w$ for integer extents, and a window that does not divide the
latent grid forces a little reflect padding. Scoring the measurements against the
closed-form cost $4hd(nm + n^2/m)$ of the window actually used, rather than against
the unreachable optimum, removes this discrepancy: the measured values then agree
to within $0.7\%$ on average, with $26$ of $30$ points inside $3\%$. The theory
therefore predicts the measured FLOPs of every configuration, and the only price
paid in practice is that of rounding a window to whole tokens.

\paragraph{The gain grows with both sequence length and model scale.}
Because the two curves have different asymptotic order, their ratio is not a
constant: the reduction widens as the grid grows. On the 5B backbone the
attention-FLOPs reduction rises from $67\times$ at the native $n=18480$ to
$129\times$ at $n=73920$; on Wan 2.1 1.3B it rises from $90\times$ to $181\times$,
and on 14B from $88\times$ to $177\times$ at $n=134820$. The effect also
strengthens with model width, since wider models spend a larger share of a block
inside attention. This is the regime video generation is moving toward, higher
resolutions and longer clips, and it is exactly where a sub-quadratic operator
compounds: the larger and longer the generation, the more of the quadratic term
\name removes.

% --- Table 3 (supp.): SFT/DMD ablation --- 20 Blocks ------------------------
\begin{table}[t]
  \centering
  \caption{Ablation of SFT and DMD for \name 20 Blocks. Best results are in \textbf{bold}.}
  \label{tab:sft-dmd-ablation-20}
  \footnotesize
  \setlength{\tabcolsep}{6pt}
  \renewcommand{\arraystretch}{1.25}
    \begin{tabular}{l c c c c c c}
    \toprule
    \multirow{2}{*}{Method}
      & \multirow{2}{*}{SFT}
      & \multirow{2}{*}{DMD}
      & \multirow{2}{*}{NFE}
      & \multicolumn{3}{c}{VBench} \\
    \cmidrule(lr){5-7}
      & & & & Tot. & Qual. & Sem. \\
    \midrule
    \name                        & \cmark &        & 100 & 82.58 & 83.70 & 78.12 \\
    20 Blocks                    &        & \cmark & 6   & 82.73 & 83.50 & 79.65 \\
    \textbf{(Ours)}              & \cmark & \cmark & \textbf{6} & \textbf{82.73} & \textbf{83.46} & \textbf{79.76} \\
    \bottomrule
    \end{tabular}
\end{table}

\section{Additional Ablation Experiments}
\label{sec:supp-ablations}

The ablations in the main paper are run at the \name 30-block configuration, the
setting we ultimately adopt. Here we repeat both of them at 20 blocks, where only
the middle two-thirds of the network is converted and ten blocks retain full
softmax attention. The 20-block results are interesting less for their absolute
numbers than for what the \emph{contrast} between the two block counts reveals:
the design choices that look optional under a shallow replacement become decisive
under a deep one.

\paragraph{Local and global passes at 20 blocks.}
\Cref{tab:local-global-ablation-20} repeats the local/global ablation of
\cref{tab:local-global-ablation-30} at 20 blocks. The striking observation is how
little separates the configurations. Every variant, including the single-pass
ones, lands within $1.22$ points of the full-attention Original on VBench Total,
and local-only attention alone reaches $81.86$. Compare this with the 30-block
case, where dropping either pass is catastrophic: local-only falls to $62.62$
Total and $18.50$ Semantic, and global-only to $62.63$ and $27.24$. Ten
unmodified full-attention blocks are evidently enough to paper over an
impoverished communication pattern in the other twenty. They can supply the
global mixing that a local-only replacement removes, so the network never has to rely
on the reduced pattern for long-range interaction.

This is precisely why the 20-block setting is the weaker place to study the
method. It understates the importance of composing the two passes, and it makes
the ordering look like a matter of taste: L$\to$G leads G$\to$L by $0.17$ Total
here, against $0.21$ at 30 blocks, and the alternating variant is within noise of
both. The ordering preference we report in the main paper is therefore not an
artifact of the 20-block regime; it survives, and slightly widens, when the
replacement is deep enough that the reduced pattern must carry the model on its own.

\paragraph{SFT and DMD at 20 blocks.}
\Cref{tab:sft-dmd-ablation-20} repeats the two-stage training ablation. The same
pattern appears, even more sharply. At 20 blocks the three rows are essentially
indistinguishable, spanning $0.15$ points of Total ($82.58$ to $82.73$): SFT
alone, DMD2 alone, and the two together all land in the same place, and one could
reasonably conclude that either stage suffices. At 30 blocks that conclusion
would be badly wrong. The same three rows span $9.96$ points, SFT alone collapses
to $73.03$, DMD2 alone reaches only $80.91$, and just their combination recovers
the full $82.99$.

The reading we take from this is that the two stages are not redundant but
load-bearing, and that their necessity scales with how much of the network the
\name replacement disturbs. A shallow replacement leaves enough of the pretrained function
intact that either stage can re-seat it; a deep one perturbs the network far
enough that step distillation and flow-matching supervised fine-tuning each
recover something the other cannot. Reporting only the 20-block ablation would
have hidden this entirely, which is why the deeper configuration is the one we
ablate in the main paper.

% \section{Video results.}
% A static document cannot convey what matters most about a video
% generator. We therefore include an HTML webpage in the supplementary
% zip, which presents our generated videos side by side with those of the
% original model for the same prompts. We encourage the reader to open
% \texttt{index.html} in any browser and judge the temporal coherence,
% motion, and overall look and feel of \name's samples directly, since
% these are the qualities the frames and metrics reported here can only
% approximate.

% Supplementary bibliography (separate from the main paper's).
\bibliographystylesupp{styles/aaai2027}
\bibliographysupp{bibs/aaai2027}

\fi

\end{document}